\documentclass[10pt,twocolumn,letterpaper]{article}

\usepackage[pagenumbers]{cvpr} 

\usepackage{xcolor}
\usepackage[table]{xcolor}
\usepackage{hhline}
\usepackage{algorithm}
\usepackage{algpseudocode}  
\usepackage{amsmath}
\usepackage{amssymb}
\usepackage{float}
\usepackage{bbding}
\usepackage{placeins}

\DeclareMathOperator*{\argmin}{arg\,min}

\usepackage{booktabs} 
\usepackage{multirow}
\usepackage{placeins}
\usepackage{subcaption}
\usepackage{lipsum} 
\usepackage{array}
\usepackage{appendix}

\definecolor{cvprblue}{rgb}{0.21,0.49,0.74}
\usepackage[pagebackref,breaklinks,colorlinks,allcolors=cvprblue]{hyperref}

\usepackage{marvosym}
\newcommand{\blfootnote}[1]{%
  \begingroup
  \renewcommand\thefootnote{}\footnote{#1}%
  \addtocounter{footnote}{-1}%
  \endgroup
}

\def\paperID{} 
\def\confName{CVPR}
\def\confYear{2026}

\title{Variation-aware Vision Token Dropping for Faster \\ Large Vision-Language Models}

\author{
\vspace{-1.2em}\\
\textbf{\large Junjie Chen$^{1*}$, Xuyang Liu$^{1*,\dagger}$, Zichen Wen$^{2}$, Yiyu Wang$^{2}$, Siteng Huang$^{3}$, Honggang Chen$^{1}$$^{\text{\Envelope}}$}\\[0.5em]
$^1$ Sichuan University, $^2$ EPIC Lab, Shanghai Jiao Tong University, $^3$ Zhejiang University\\
\textbf{Code:} \url{https://github.com/xuyang-liu16/V2Drop}
}

\begin{document}
\maketitle
\blfootnote{$*$ Equal contributions. $^{\text{\Envelope}}$ Corresponding author: Honggang Chen (\texttt{honggang\_chen@scu.edu.cn}). $\dagger$ Project leader: Xuyang Liu (\texttt{seanleo666@gmail.com}).}

\begin{abstract}

Large vision-language models (LVLMs) have demonstrated remarkable capabilities in multimodal understanding tasks. However, the increasing demand for high-resolution image and long-video understanding results in substantial token counts, consequently leading to reduced inference efficiency. Token compression offers a direct solution by reducing the number of tokens to be processed, thereby improving computational efficiency without architectural changes. Through extensive analysis, we identify two critical limitations in existing inner-LLM token compression methods: positional bias and incompatibility with efficient operators, which critically hinder their practical deployment for LVLM acceleration. This paper presents the first approach from a dynamic token variation perspective, revealing that visual token variations within LLMs exhibit task-agnostic properties. We propose Variation-aware Vision Token Dropping (\textit{i.e.}, \textbf{V$^2$Drop}), which progressively removes visual tokens with minimal variation during LVLM inference, thereby enhancing computational efficiency. Extensive experiments across multiple models and benchmarks consistently demonstrate that V$^2$Drop maintains \textbf{94.0\%} and \textbf{98.6\%} of the original performance for image and video understanding tasks respectively, while reducing LLM generation latency by \textbf{31.5\%} and \textbf{74.2\%}.

\end{abstract}

\section{Introduction}
\label{sec:introduction}

Large vision-language models (LVLMs) have demonstrated remarkable capabilities in visual understanding and reasoning~\cite{Liu:LLaVA-1.5,Wang:Qwen2-VL,li2024llava-ov,wen2026innovator}, excelling across diverse vision-language tasks. However, high-resolution image understanding~\cite{Chen:InternVL-v1.5,guo2025seed1} and long video comprehension~\cite{chen2024longvila,zhang2024llava-video} introduce substantial visual tokens that significantly reduce computational efficiency and hinder practical deployment~\cite{liu2025shifting}.

To address this challenge, numerous token compression methods have emerged to eliminate redundant visual tokens, enhancing LVLM efficiency while preserving performance~\cite{Chen:FastV,arif2024hired,wen2025dart,chen2025ipcv}. Methods that compress visual tokens within LLMs have gained particular attention due to their architecture-agnostic and plug-and-play nature~\cite{Zhang:SparseVLM,xing2024pdrop,Han2024:FiCoCo,liu2025mixkv}. These approaches leverage LLM attention weights to selectively retain important visual tokens while pruning less significant ones, thereby accelerating inference. However, despite promising results, these methods face fundamental challenges that limit their practical deployment.

\begin{figure}[!t]
    \centering
    \includegraphics[width=\linewidth]{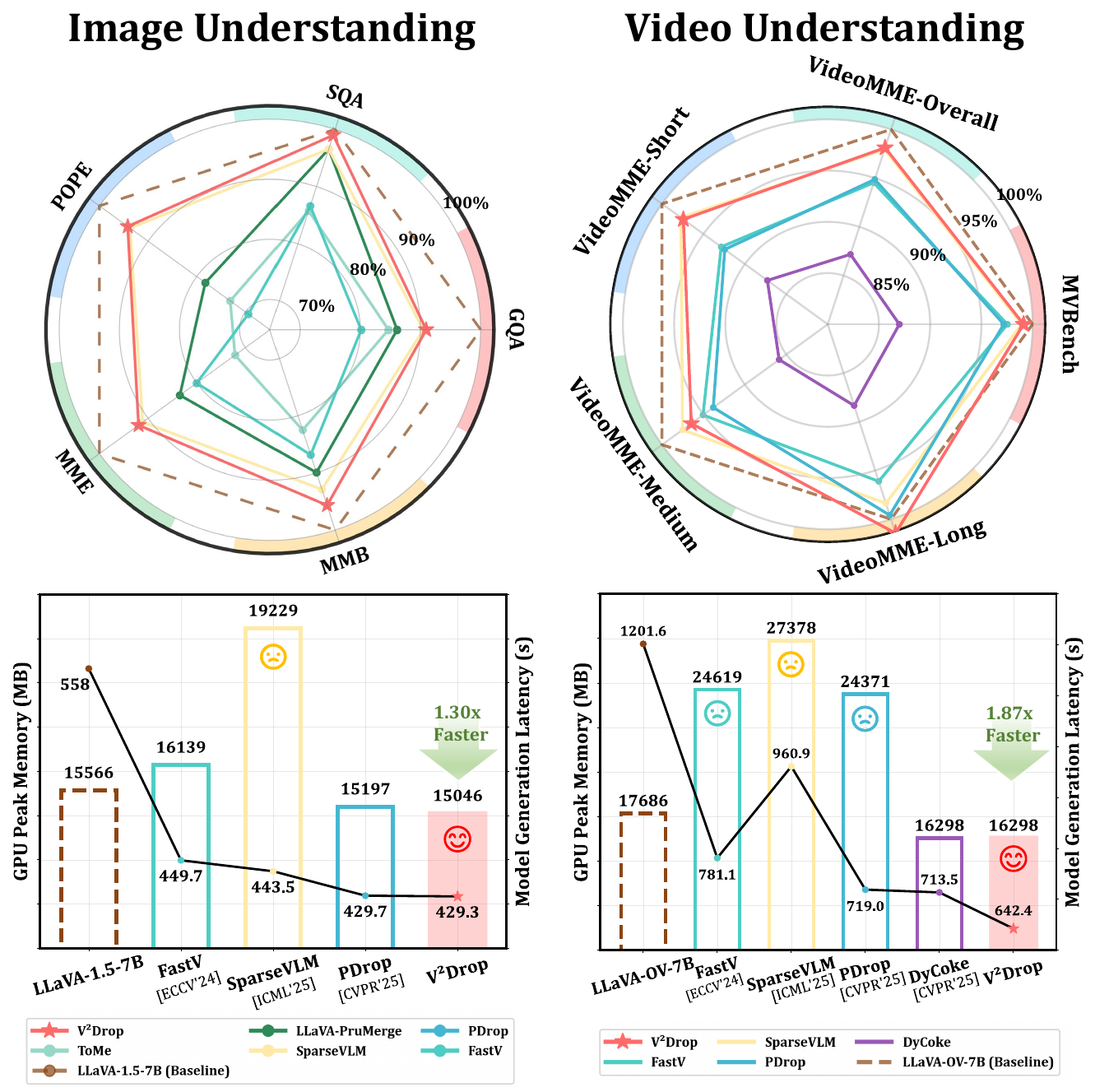}
    \vspace{-6mm}
    \caption{\textbf{Performance-Efficiency trade-offs comparison.} V$^2$Drop achieves superior performance-efficiency trade-offs across both image and video understanding tasks.}
    \vspace{-4mm}
\label{fig:performance_efficiency}
\end{figure}

\begin{figure*}[t]
    \centering
    \includegraphics[width=\textwidth]{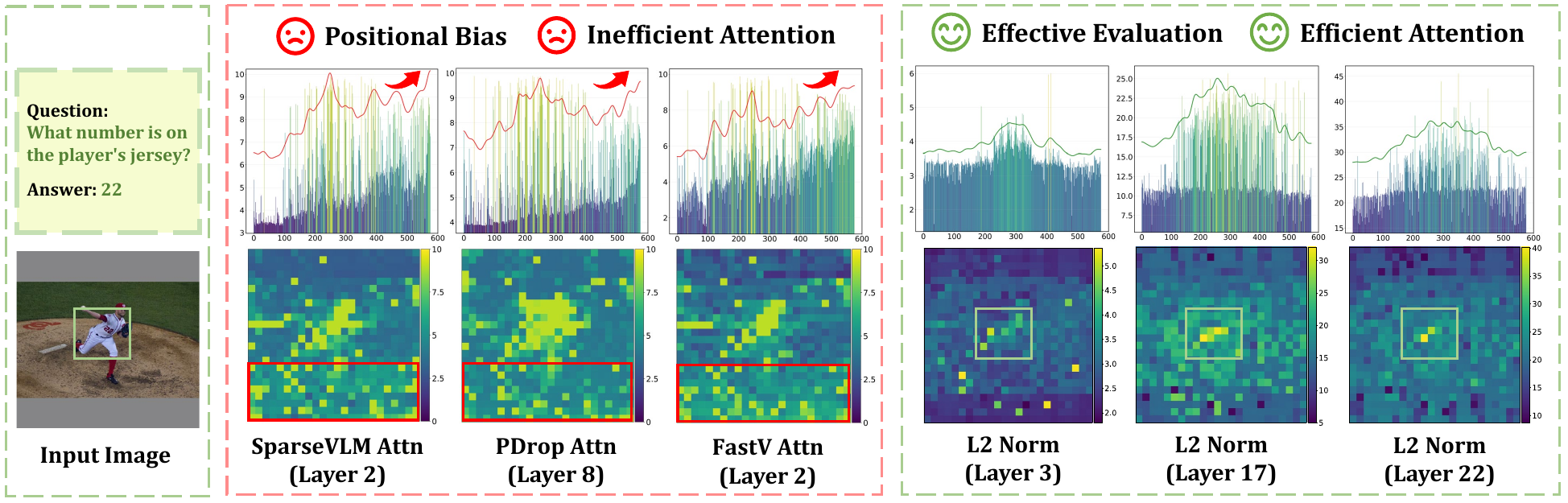}
    \vspace{-6mm}
    \caption{\textbf{Attention-guided token evaluation vs. variation-aware token evaluation.} Attention-guided methods (\textit{e.g.}, FastV~\cite{Chen:FastV}, PDrop~\cite{xing2024pdrop}, SparseVLM~\cite{Zhang:SparseVLM}) exhibit \textit{information-agnostic positional bias}, assigning high importance to later positions regardless of content (red arrows and boxes), and are \textit{incompatible with efficient operators}. In contrast, measuring token-wise variation  (\textit{e.g.}, L2 Norm) intuitively reflects token importance (green boxes) while maintaining compatibility with efficient operators. The red and green curves show the trends of attention scores and variation scores, respectively.}
    \vspace{-4mm}
    \label{fig:position_bias}
\end{figure*}

\begin{figure}[!t]
    \centering
    \includegraphics[width=\linewidth]{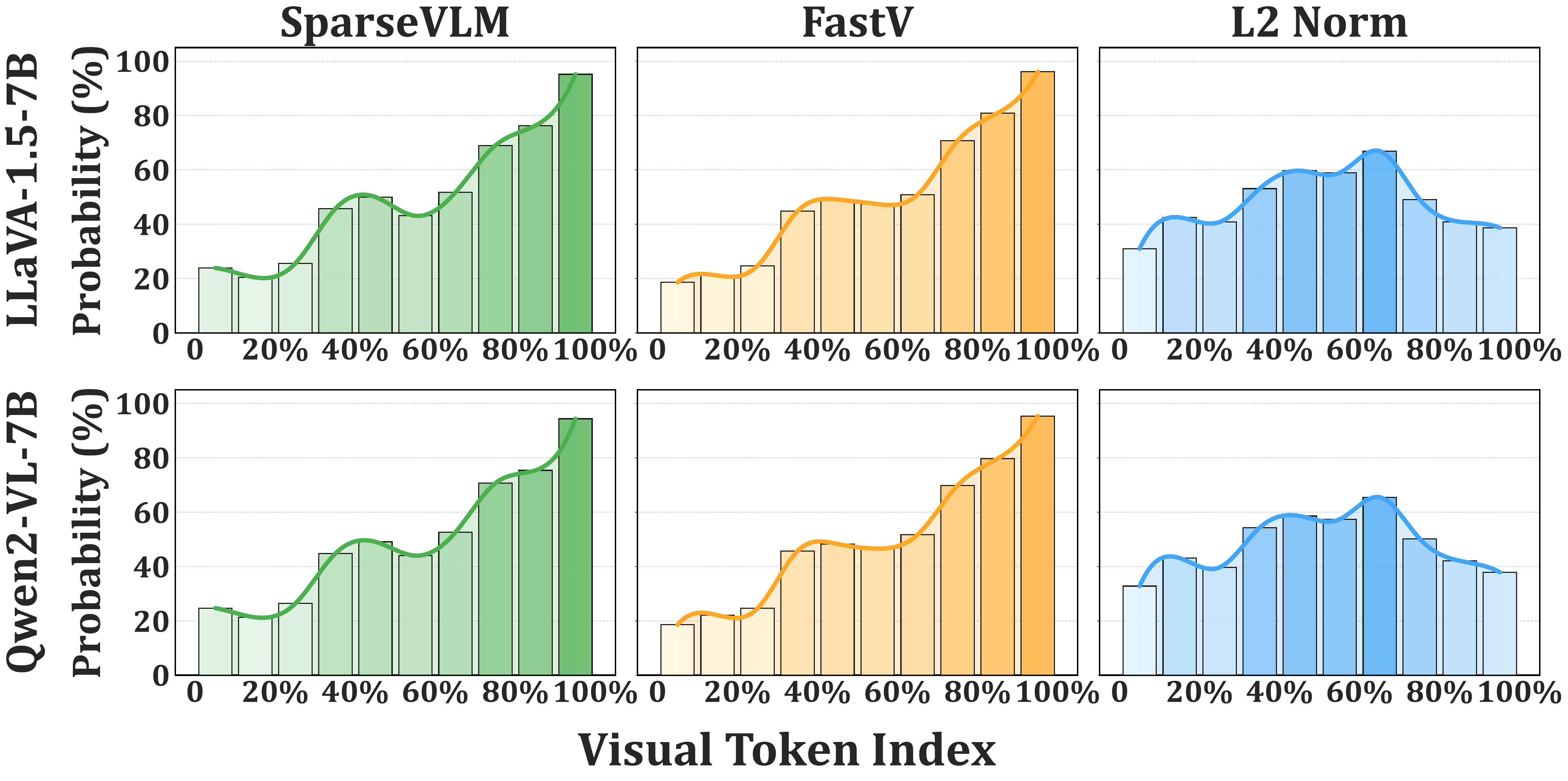}
    \vspace{-6mm}
    \caption{\textbf{Analysis of the distribution of retained visual tokens with respect to token index positions.} Visual tokens with larger indexes are located at the bottom of images. The probability denotes the chance of each token being retained during pruning.} 

    \vspace{-4.5mm}
\label{fig:position_bias_quantity}
\end{figure}

To investigate this phenomenon, we analyzed visual tokens retained by attention-guided methods (\textit{e.g.}, FastV~\cite{Chen:FastV}, SparseVLM~\cite{Zhang:SparseVLM}) and variation-aware evaluation (L2 Norm) using LLaVA-1.5-7B and Qwen2-VL-7B on TextVQA, POPE, and MME. Figure~\ref{fig:position_bias} illustrates this through a representative example, while Figure~\ref{fig:position_bias_quantity} provides quantitative evidence: attention-guided methods exhibit strong bias toward retaining tokens at the end of visual sequences regardless of content, while variation-aware evaluation produces naturally uniform spatial distributions.

This analysis reveals that existing attention-guided compression methods~\cite{xing2024pdrop,Zhang:SparseVLM} suffer from a fundamental flaw: \textit{relying on external attention signals rather than intrinsic token properties}, leading to \textbf{\textit{two critical issues:}} \textbf{(i) Information-agnostic positional bias:} These methods systematically assign high importance to later-positioned tokens regardless of visual content, retaining irrelevant information while discarding important tokens, thereby exacerbating multi-modal hallucinations. \textbf{(ii) Incompatibility with efficient operators:} Computing attention weights conflicts with efficient mechanisms attention (\textit{e.g.}, FlashAttention~\cite{daoFlashAttention-2}), resulting in peak memory usage exceeding uncompressed models (Table~\ref{tab:efficiency_comparison_merged}). This raises a fundamental question: \textit{Instead of relying on indirect attention signals, can we directly assess token importance through their intrinsic behavioral patterns within the model?}.

To address this fundamental challenge, we propose a paradigm shift from external signal dependence to intrinsic property analysis. We present the \textbf{first} study from a \textbf{\textit{token variation}} perspective, revealing that visual token variations within LLMs naturally encode importance information. Our key insight is that \textit{lazy tokens}—those showing minimal variation across LLM layers—less likely to impact final predictions and can be safely removed without performance degradation. Based on this finding, we propose \textbf{V}ariation-aware \textbf{V}ision Token \textbf{Drop}ping (\textit{i.e.}, \textbf{V$^2$Drop}), which naturally avoids positional bias by focusing on intrinsic token dynamics and maintains full compatibility with efficient operators by eliminating attention weight computation.

Consequently, V$^2$Drop demonstrates outstanding performance and efficiency across both image and video understanding tasks, achieving \textbf{1.30$\times$} and \textbf{1.87$\times$} acceleration for image and video understanding, as shown in Figure~\ref{fig:performance_efficiency}. The main contributions are summarized as follows:
\begin{itemize}
    \item \textbf{Systematic Analysis of Token Variation Patterns:} We conduct the first comprehensive analysis of visual token evolution within LVLMs, revealing that token-wise variation magnitudes correlate with task relevance and effectively reflect token importance, pioneering token compression from the variation perspective.
    \item \textbf{Variation-aware Token Dropping:} We propose V$^2$Drop, a variation-aware compression method that identifies and progressively drops tokens based on their intrinsic behavioral patterns, eliminating positional bias while maintaining compatibility with efficient operators.
    \item \textbf{Comprehensive Performance-Efficiency Trade-offs:} Extensive experiments demonstrate that V$^2$Drop achieves exceptional performance across diverse LVLMs and VideoLLMs, with comprehensive analyses validating its robustness in balancing accuracy and efficiency.
\end{itemize}

\begin{figure}[!t]
    \centering
    \includegraphics[width=\linewidth]{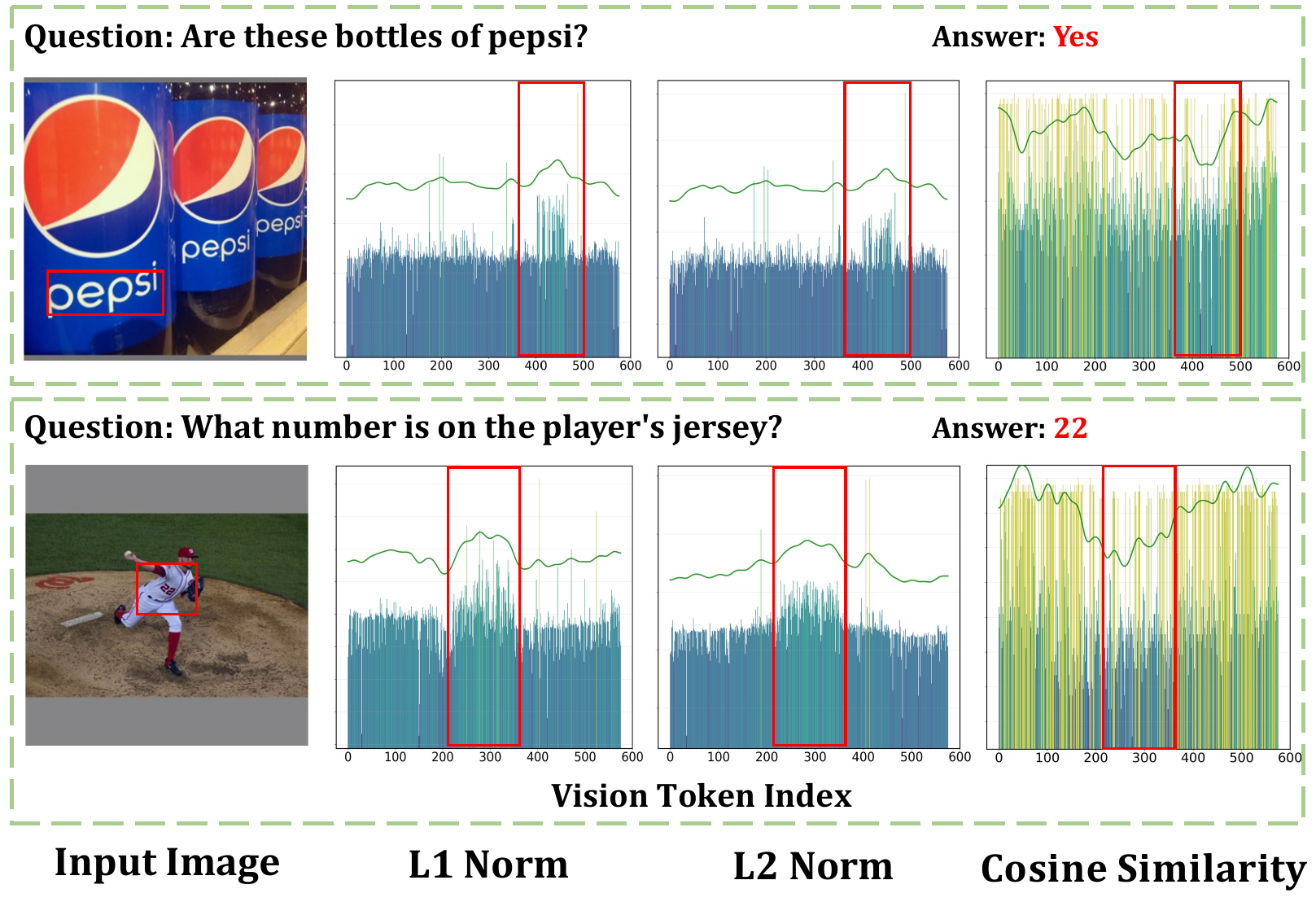}
    \vspace{-6mm}
    \caption{\textbf{Quantifying vision token variation with different metrics.} Regions corresponding to the answer exhibit significant variation magnitudes (red boxes). The green curve shows the trends of variation scores.}
    \vspace{-4mm}
\label{fig:token_variation}
\end{figure}

\section{Related Work}
\label{sec:related_work}

\noindent \textbf{Large Vision-Language Models.}
Large vision-language models (LVLMs) integrate a vision encoder (\textit{i.e.}, ViT), projection module, and LLM for multi-modal comprehension~\cite{liu2023llava,Dai:InstructBLIP,Chen:InternVL-v1.5,Bai:Qwen-VL}. Recent work introduces higher-resolution inputs through dynamic cropping (\textit{e.g.}, InternVL-3~\cite{zhu2025internvl3}, LLaVA-OneVision~\cite{li2024llava-ov}) and native resolution methods (\textit{e.g.}, Qwen2-VL~\cite{Wang:Qwen2-VL}, Seed1.5-VL~\cite{guo2025seed1}). Similarly, VideoLLMs process increasingly longer sequences (\textit{e.g.}, LLaVA-Video~\cite{zhang2024llava-video}, VideoLLaMA3~\cite{zhang2025videollama}), with VideoXL-Pro~\cite{liu2025video} achieving multi-hour frame-level understanding. However, this substantially increases visual tokens, introducing quadratic computational complexity, severely constraining scalability and practical deployment.

\vspace{0.3em}
\noindent \textbf{Token Compression for LVLMs.} Token compression directly reduces sequence length to improve model efficiency~\cite{liu2025shifting,shao2026survey}, evolving from training-aware paradigms~\cite{Li:TokenPacker,wen2025epic} to training-free methods~\cite{Chen:FastV} that enable plug-and-play LVLM inference acceleration. They can be categorized into two paradigms: \textbf{(i)} Pre-LLM compression~\cite{Shang:LLaVA-PruMerge,Yang2024:Visionzip,liu2025globalcom2}, which compresses visual tokens before LLM, and \textbf{(ii)} Inner-LLM compression~\cite{Zhang:SparseVLM,xing2024pdrop}, which performs compression during LLM forward propagation. However, most Inner-LLM methods rely on attention weights, making them incompatible with efficient operators like FlashAttention~\cite{daoFlashAttention-2} and causing substantial memory increases for VideoLLMs~\cite{liu2025vidcom2,shao2025holitom,wang2025stc,tao2024dycoke}. Additionally, they exhibit positional bias, favoring tokens near final positions regardless of semantic relevance~\cite{wen2025token,wen2025dart,liu2025globalcom2}.

In this work, we explore progressively dropping low-variation vision tokens during LLM inference, maintaining compatibility with efficient operators while eliminating positional bias for training-free LVLM acceleration.

\begin{figure*}[t]
    \centering
    \includegraphics[width=\textwidth]{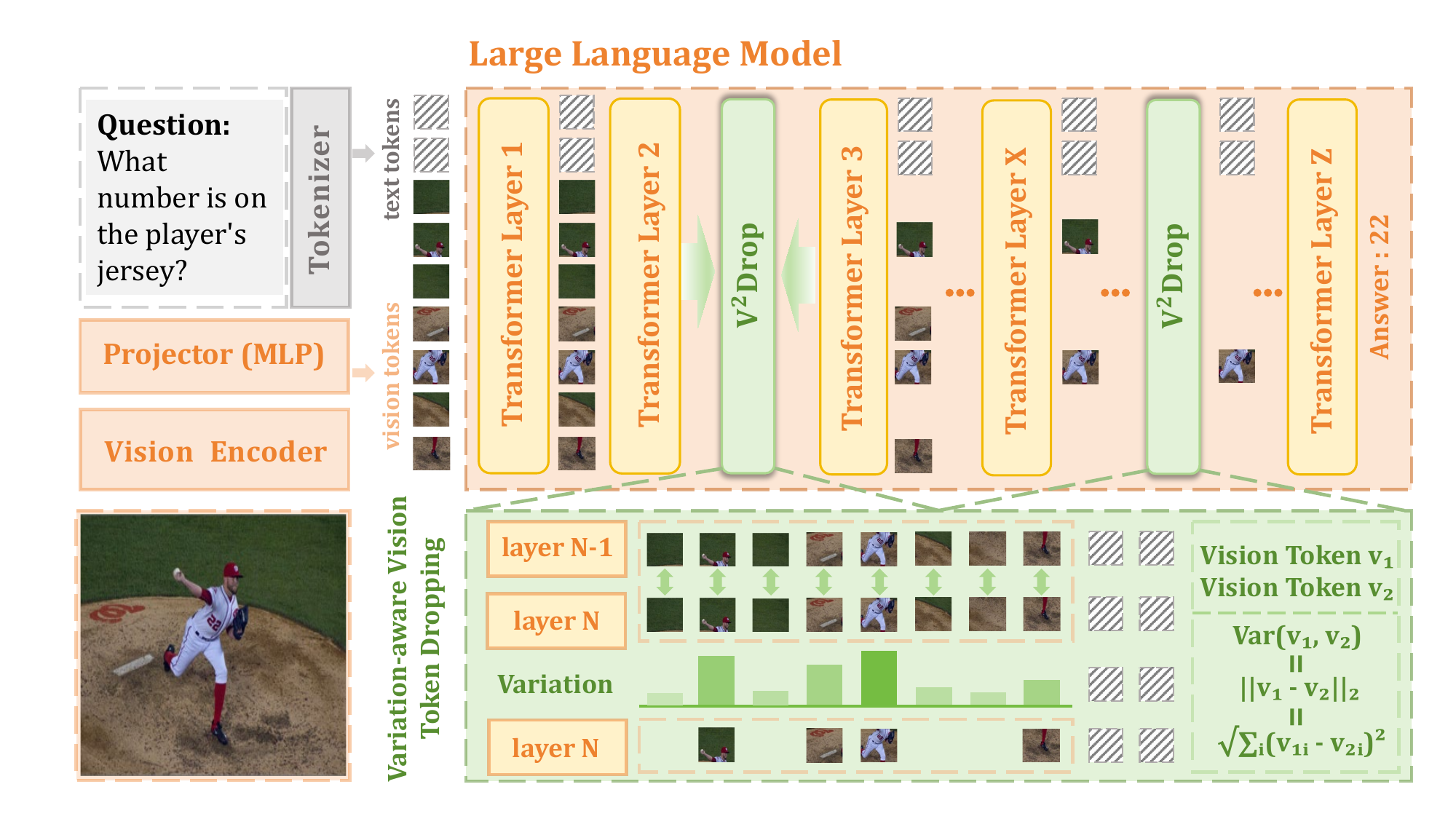}
    \vspace{-6mm}
    \caption{\textbf{Overall framework of V$^2$Drop.} V$^2$Drop measures token-wise variation across adjacent LLM layers and progressively drops vision tokens with minimal variation (\textit{i.e.}, lazy tokens), thereby achieving plug-and-play LVLM inference acceleration.} 
    \vspace{-4mm}
    \label{fig:overview}
\end{figure*}

\section{Methodology}
\label{sec:method}

\subsection{Preliminary: LVLMs}
\label{subsec:Preliminary}

\noindent \textbf{LVLM Architecture.} Most current LVLMs adopt the ``ViT-Projector-LLM'' paradigm~\cite{Liu:LLaVA-1.5,Wang:Qwen2-VL} comprising three key components. Given an image $\mathbf{I} \in \mathbb{R}^{H \times W \times 3}$ or video $\mathbf{V} = \left\{ \mathbf{v}_i \right\}_{i=1}^T \in \mathbb{R}^{T \times H \times W \times 3}$: \textbf{(i)} A visual encoder (ViT) first encodes the input into visual embeddings $\mathbf{E} \in \mathbb{R}^{N \times D}$ for images or $\mathbf{E} = \left\{ \mathbf{e}_i \right\}_{i=1}^T \in \mathbb{R}^{T \times N \times D}$ for videos; \textbf{(ii)} A projector (typically a 2-layer MLP) transforms these embeddings into vision tokens $\mathbf{F}^v \in \mathbb{R}^{M \times D'}$ for images or $\mathbf{F}^v = \left\{ \mathbf{f}_i^v \right\}_{i=1}^T \in \mathbb{R}^{T \times M \times D'}$ for videos, where $M \leq N$\footnote{Some LVLMs like LLaVA-1.5 maintain the same number of tokens ($M = N$) during projection, while others, particularly VideoLLMs, reduce the token count ($M < N$) for efficiency.}; and \textbf{(iii)} An LLM decoder processes all visual and textual tokens $\mathbf{F}^t$ during prefilling, then autoregressively generates response tokens during decoding:
{\setlength\abovedisplayskip{2mm}
\setlength\belowdisplayskip{2mm}
\begin{equation}
p\left(\mathbf{Y} \mid \mathbf{F}^{v}, \mathbf{F}^{t}\right)=\prod_{j=1}^L p\left(\mathbf{y}_j \mid \mathbf{F}^{v}, \mathbf{F}^{t}, \mathbf{Y}_{1:j-1}\right),
\end{equation}
}
where $\mathbf{Y}=\left\{ \mathbf{y}_j \right\}_{j=1}^L$ denotes the generated tokens.


\subsection{Token Variation in LVLMs}
\label{subsec:Token_Variation}

To overcome attention-guided limitations, we shift from external signals to intrinsic token properties. These fundamental limitations prompt us to reconsider the essence of token importance: \textit{what makes a vision token genuinely crucial for LLM visual understanding?} Our key insight is that: \textbf{\textit{tokens genuinely participating in LLM computation exhibit significant representational changes across layers, while less important tokens remain relatively static}}. We analyze the relationship between token variation magnitude and token importance to validate this hypothesis.

\vspace{0.3em}
\noindent \textbf{Token Variation Metrics.} We measure token variation between consecutive LLM transformer layers using three metrics (\textit{i.e.}, L1 Distance, L2 Distance, and Cosine Similarity):

{\setlength\abovedisplayskip{2mm}
\setlength\belowdisplayskip{2mm}
\begin{equation}
\small
\text{Var}(\mathbf{f}_i^{(l-1)}, \mathbf{f}_i^{(l)}) = 
\begin{cases}
\|\mathbf{f}_i^{(l)} - \mathbf{f}_i^{(l-1)}\|_1 & \text{(L1 Distance)} \\
\|\mathbf{f}_i^{(l)} - \mathbf{f}_i^{(l-1)}\|_2 & \text{(L2 Distance)} \\
1 - \frac{\mathbf{f}_i^{(l)} \cdot \mathbf{f}_i^{(l-1)}}{\|\mathbf{f}_i^{(l)}\|_2 \|\mathbf{f}_i^{(l-1)}\|_2} & \text{(Similarity)},
\end{cases}
\end{equation}
}
where $\mathbf{f}_i^{(l)}$ denotes the $i$-th vision token at layer $l$, and higher variation values indicate greater representational changes. Each metric captures distinct transformation patterns: L1 distance measures sparse changes, L2 distance captures overall magnitude, and cosine similarity reflects directional changes in the representation space.

\vspace{0.3em}
\noindent \textbf{Variation-Relevance Relationship.} Figure~\ref{fig:token_variation} demonstrates token variation quantification across three metrics for visual tokens in the third transformer layer of LLaVA-1.5-7B. We observe a \textbf{consistent and crucial pattern}: tokens exhibiting significant variation (high L1/L2 distances, low cosine similarity) consistently correspond to question-relevant regions (red boxes), encoding rich semantic information essential for task completion. Conversely, tokens with minimal variation—termed \textit{lazy tokens}—correspond to task-irrelevant regions with limited impact to final predictions

Importantly, Figure~\ref{fig:token_variation} presents two distinct cases where question-relevant regions appear in different spatial locations (bottom and center), corresponding to middle and posterior token positions. \textbf{All three variation metrics accurately capture these semantically important regions regardless of spatial position}, demonstrating the robustness of using token variation to measure token importance against positional bias. This spatial-agnostic detection capability represents a fundamental advantage over attention-guided approaches that suffer from positional bias.

These findings validate our core hypothesis: high-variation vision tokens actively participate in reasoning processes, encoding semantically crucial information that must be preserved for optimal performance. Conversely, lazy tokens maintain stable representations throughout LLM processing, indicating limited contribution to final predictions.

\subsection{Variation-aware Vision Token Dropping}
\label{subsec:V2Drop}

Building on this key insight, we propose \textbf{V}ariation-aware \textbf{V}ision Token \textbf{Drop}ping (\textbf{V$^2$Drop}), a novel approach that progressively identify and efficiently drop lazy tokens by measuring vision token variation magnitudes in LLMs while preserving semantically important ones.

Given a sequence of vision tokens $\mathbf{F}^v \in \mathbb{R}^{M \times D'}$ in the LLM, we adopt a \textit{multi-stage progressive dropping} strategy as illustrated in Figure~\ref{fig:overview}. We perform pruning at three strategically selected layers $\mathcal{L}$ spanning shallow, middle, and deep stages of the LLM to balance compression efficiency and performance preservation across model depth. At each pruning layer $l_k \in \mathcal{L}$, our framework performs three key operations:

\noindent \textbf{(i) Variation Computation:} For each vision token $\mathbf{f}_i^{(l_k)}$ at layer $l_k$, we compute variation scores by measuring representational changes from the previous layer $l_k-1$:
{\setlength\abovedisplayskip{2mm}
\setlength\belowdisplayskip{2mm}
\begin{equation}
\mathbf{S}^{(l_k)} = \{\text{Var}(\mathbf{f}_i^{(l_k-1)}, \mathbf{f}_i^{(l_k)})\}_{i=1}^{M_{l_k}},
\end{equation}
}
where $M_{l_k}$ is the number of vision tokens at layer $l_k$, where we empirically use L2 distance by default (ablation study of variation metrics is in Figure~\ref{fig:ablation1}). This efficiently captures token evolution without attention weight re-computation.

\noindent \textbf{(ii) Token Ranking and Selection:} Vision tokens are ranked by variation scores in descending order, and we retain the top-$K_{l_k}$ tokens with highest scores:
{\setlength\abovedisplayskip{2mm}
\setlength\belowdisplayskip{2mm}
\begin{equation}
\mathbf{\hat{F}}^v_{l_k} = \text{TopK}(\mathbf{F}^v_{l_k}, \mathbf{S}^{(l_k)}, K_{l_k}).
\end{equation}
}
This naturally filters out lazy tokens while preserving semantically important ones, avoiding positional bias.

\noindent \textbf{(iii) Token Reorganization:} The selected vision tokens are reorganized for subsequent layers, where $K_{l_k} < M_{l_k}$ ensures progressive visual token dropping.

The dropping process follows a pre-defined schedule:
{\setlength\abovedisplayskip{2mm}
\setlength\belowdisplayskip{2mm}

\begin{equation}
M \rightarrow K_a \rightarrow K_{b} \rightarrow K_{c},
\end{equation}
}
where $K_a$, $K_{b}$, and $K_{c}$ are predefined compression targets adjustable for performance-efficiency trade-offs. Ablation studies show that V$^2$Drop is robust to layer selection (Figure~\ref{fig:ablation2}) and that progressive dropping significantly outperforms one-time dropping (Figure~\ref{fig:ablation3}). Detailed experimental configurations are provided in the Appendix.

\subsection{Theoretical Analysis}
To validate our variation-aware token dropping strategy, we establish a theoretical connection between token variation magnitude and model output through first-order analysis.

\subsubsection{Problem Formulation}
Let $X^{(t)} = \{x_1^{(t)}, \ldots, x_n^{(t)}\} \subset \mathbb{R}^d$ denote token representations at layer $t$. We define the \textbf{inter-layer variation} of token $j$ as:
\begin{align}
\Delta x_j^{(t)} = x_j^{(t+1)} - x_j^{(t)}.
\end{align}
Let $f: \mathbb{R}^{n \times d} \to \mathbb{R}^k$ map layer $(t+1)$ representations to the final output. The Jacobian with respect to token $j$ is:
\begin{align}
J_j = \frac{\partial f}{\partial x_j^{(t+1)}} \in \mathbb{R}^{k \times d}.
\end{align}

\subsubsection{Variation-Impact Theorem}
\textbf{Theorem 1.} Under mild smoothness assumptions on $f$, the output change induced by token $j$ satisfies:
\begin{align}
\| \Delta f_j \| \approx \|J_j\|_{\text{op}} \cdot \|\Delta x_j^{(t)}\|,
\end{align}
where $\Delta f_j$ denotes the output change when only token $j$ varies from layer $t$ to $t+1$, and $\|\cdot\|_{\text{op}}$ is the operator norm.

\textbf{Proof.} By first-order Taylor expansion around $x_j^{(t)}$:
\begin{align}
f(\ldots, x_j^{(t+1)}, \ldots) 
&= f_j^{(t)} + J_j \, \Delta x_j^{(t)} + \mathcal{O}(\|\Delta x_j^{(t)}\|^2) \nonumber \\
&= f_j^{(t)} + J_j \, \Delta x_j^{(t)} + R_j,
\end{align}
where \( f_j^{(t)} = f(\ldots, x_j^{(t)}, \ldots) \) and $R_j$ denotes the higher-order remainder term satisfying $\|R_j\| = \mathcal{O}(\|\Delta x_j^{(t)}\|^2)$.

Taking norms on both sides and applying the triangle inequality:
\begin{align}
\|\Delta f_j\| 
&= \|f(\ldots, x_j^{(t+1)}, \ldots)
   - f(\ldots, x_j^{(t)}, \ldots)\| \nonumber \\
&\leq \|J_j \cdot \Delta x_j^{(t)}\|
   + \|R_j\|.
\end{align}

By the definition of operator norm:
\begin{align}
\|J_j \cdot \Delta x_j^{(t)}\| \leq \|J_j\|_{\text{op}} \cdot \|\Delta x_j^{(t)}\|.
\end{align}
For sufficiently small $\|\Delta x_j^{(t)}\|$ (typically satisfied in deep networks with residual connections where layer-wise changes are bounded), the quadratic term $\|R_j\|$ is negligible compared to the linear term. Thus:
\begin{align}
\|\Delta f_j\| \approx \|J_j\|_{\text{op}} \cdot \|\Delta x_j^{(t)}\|.
\end{align}
(Appendix for full proof.)\hfill $\square$

\subsubsection{Implications}
\textbf{Corollary.} Under the assumptions of Theorem 1, larger variation 
$\|\Delta x_j^{(t)}\|$ implies greater output influence, providing a 
computationally efficient proxy for token importance.

\textbf{Proof.} For tokens with $\|J_j\|_{\text{op}} \geq \mu > 0$, 
substituting into the theorem yields:
\begin{align}
\|\Delta f_j\| \gtrsim \mu \cdot \|\Delta x_j^{(t)}\|.
\end{align}
Therefore, tokens with larger $\|\Delta x_j^{(t)}\|$ induce proportionally 
larger output changes. (Appendix for full proof.)\hfill $\square$

\section{Experiments}
\label{sec:Experiments}

\noindent \textbf{Experiment Setting.} 
We conduct comprehensive on various LVLMs and VideoLLMs across ten diverse benchmarks, with implementation details in the Appendix.

\vspace{0.5em}
\noindent \textbf{Computational Overhead.} We prune at three layers. Computing L2 distances for $M$ tokens of dimension $D'$ requires $3MD'$ FLOPs ($\sim$7M for $M{=}576$, $D'{=}4096$ in LLaVA-1.5), only 0.022\% of a single attention layer (32B FLOPs). The total overhead across three layers ($\sim$21M FLOPs) is merely 0.002\% of the full forward pass. Table~\ref{tab:efficiency_comparison_merged} confirms this negligible cost: V$^2$Drop and random dropping achieve nearly identical throughput (9.01 vs 9.08 items/s).

\begin{table*}[!t]
    \vspace{3mm}

    \centering
    \setlength\tabcolsep{11pt}  
    \setlength{\arrayrulewidth}{0pt}
    \renewcommand{\arraystretch}{0.6}  
    {\fontsize{10pt}{10pt}\selectfont  

    \definecolor{softgreen}{HTML}{66BB6A}   
    \definecolor{softred}{HTML}{E85C5C}     
    \definecolor{softgray}{HTML}{A6A6A6}    
    
    \newcommand{\downtiny}[1]{{\!\scriptsize{#1}}}
    \newcommand{\greentiny}[1]{\textcolor{softgreen}{#1}}
    \newcommand{\redtiny}[1]{\textcolor{softred}{#1}}
    \newcommand{\graytiny}[1]{\textcolor{softgray}{#1}}
    \begin{tabular}{lccccccc}
    \toprule
    \textbf{Methods} & \textbf{GQA} & \textbf{SQA} & \textbf{TextVQA} & \textbf{POPE} & \textbf{MME} & \textbf{MMBench} & \textbf{Average} \\
    \midrule
    \multicolumn{8}{l}{\textit{Upper Bound, 576 Tokens (100\%)}} \\
    LLaVA-1.5-7B~\cite{Liu:LLaVA-W} & \graytiny{61.9} & \graytiny{69.5} & \graytiny{58.2} & \graytiny{85.9} & \graytiny{1862} & \graytiny{64.6} & \graytiny{100.0\%} \\
    \midrule
    \multicolumn{8}{c}{\textit{Average Retain 192 Tokens ($\downarrow$66.7\%)}} \\
    \midrule
    ToMe \downtiny{[ICLR'23]} & 54.3 & 65.2 & 52.1 & 72.4 & 1563 & 60.5 & 88.8\% \\
    FastV \downtiny{[ECCV'24]} & 52.7 & 67.3 & 52.5 & 64.8 & 1612 & 61.2 & 88.2\% \\
    HiRED \downtiny{[AAAI'25]} & \textbf{58.7} & 68.4 & 47.4 & 82.8 & 1737 & 62.8 & 93.6\% \\
    LLaVA-PruMerge \downtiny{[ICCV'25]} & 54.3 & 67.9 & 54.3 & 71.3 & 1632 & 59.6 & 90.3\% \\
    SparseVLM \downtiny{[ICML'25]} & 57.6 & 69.1 & \textbf{56.1} & 83.6 & 1721 & 62.5 & 95.9\% \\
    PDrop \downtiny{[CVPR'25]} & 57.1 & 68.8 & \textbf{56.1} & 82.3 & 1766 & 63.2 & 96.0\% \\
    \rowcolor{gray!15}
    \textbf{V$^2$Drop} & 58.5 & \textbf{69.3} & 55.6 & \textbf{85.1} & \textbf{1826} & \textbf{63.7} & \textbf{97.6\%} \\
    \midrule
    \multicolumn{8}{c}{\textit{Average Retain 128 Tokens ($\downarrow$77.8\%)}} \\
    \midrule
    ToMe \downtiny{[ICLR'23]} & 52.4 & 59.6 & 49.1 & 62.8 & 1343 & 53.3 & 80.4\% \\
    FastV \downtiny{[ECCV'24]} & 49.6 & 60.2 & 50.6 & 59.6 & 1490 & 56.1 & 81.7\% \\
    HiRED \downtiny{[AAAI'25]} & \textbf{57.2} & 68.1 & 46.1 & 79.8 & 1710 & 61.5 & 91.6\% \\
    LLaVA-PruMerge \downtiny{[ICCV'25]} & 53.3 & 67.1 & 54.3 & 67.2 & 1554 & 58.1 & 87.9\% \\
    SparseVLM \downtiny{[ICML'25]} & 56.0 & 67.1 & \textbf{54.9} & 80.5 & 1696 & 60.0 & 93.2\% \\
    PDrop \downtiny{[CVPR'25]} & 56.0 & 68.3 & 54.8 & \textbf{82.3} & 1644 & 61.1 & 93.6\% \\
    \rowcolor{gray!15}
    \textbf{V$^2$Drop} & 56.3 & \textbf{68.8} & 53.8 & 80.9 & \textbf{1712} & \textbf{61.8} & \textbf{94.0\%} \\
    \midrule
    \multicolumn{8}{c}{\textit{Average Retain 64 Tokens ($\downarrow$88.9\%)}} \\
    \midrule
    ToMe \downtiny{[ICLR'23]} & 48.6 & 50.0 & 45.3 & 52.5 & 1138 & 43.7 & 69.7\% \\
    FastV \downtiny{[ECCV'24]} & 46.1 & 51.1 & 47.8 & 48.0 & 1256 & 48.0 & 71.3\% \\
    LLaVA-PruMerge \downtiny{[ICCV'25]} & 51.9 & 68.1 & 54.0 & 65.3 & \textbf{1549} & 55.2 & 86.5\% \\
    SparseVLM \downtiny{[ICML'25]} & \textbf{52.7} & 62.2 & \textbf{51.8} & \textbf{75.1} & 1505 & \textbf{56.2} & 86.5\% \\
    PDrop \downtiny{[CVPR'25]} & 41.9 & 68.6 & 45.9 & 55.9 & 1092 & 33.3 & 70.1\% \\
    \rowcolor{gray!15}
    \textbf{V$^2$Drop} & 50.5 & \textbf{68.9} & \textbf{51.8} & \textbf{75.1} & 1470 & 55.2 & \textbf{86.9\%} \\
    \bottomrule
    \end{tabular}
    }
    \vspace{-2mm}
    \captionof{table}{\textbf{Comparison with other token compression methods with LLaVA-1.5-7B across image understanding benchmarks.} ``Average'' shows the mean performance across benchmarks at different retention ratios, with \textbf{best} results highlighted.}
    \vspace{-3mm}
    \label{tab:image_comparison_results_1}
\end{table*}

\begin{table}[t!]

    \centering
    \setlength\tabcolsep{2.2pt}
    \renewcommand{\arraystretch}{1.0}
    \vspace{0pt}
    {\fontsize{7.5pt}{7.5pt}\selectfont  
    
    \definecolor{softgreen}{HTML}{66BB6A}   
    \definecolor{softred}{HTML}{E85C5C}     
    \definecolor{softgray}{HTML}{A6A6A6}    
    
    \newcommand{\downtiny}[1]{{\!\scriptsize{#1}}}
    \newcommand{\greentiny}[1]{\textcolor{softgreen}{#1}}
    \newcommand{\redtiny}[1]{\textcolor{softred}{#1}}
    \newcommand{\graytiny}[1]{\textcolor{softgray}{#1}}
    
    \begin{tabular}{lccccccc}
    \toprule
    \textbf{Methods} & \textbf{AI2D} & \textbf{MMStar} & \textbf{SQA} & \textbf{POPE} & \textbf{MME} & \textbf{MMB} & \textbf{Avg.} \\
    \midrule
    \multicolumn{8}{l}{\textit{Upper Bound, All Tokens (100\%)}} \\
    Qwen2-VL-7B~\cite{Wang:Qwen2-VL} & \graytiny{82.1} & \graytiny{59.6} & \graytiny{84.7} & \graytiny{86.1} & \graytiny{2317} & \graytiny{80.5} & \graytiny{100.0\%} \\
    \midrule
    \multicolumn{8}{c}{\textit{Token Reduction ($\downarrow$66.7\%)}} \\
    \midrule
    FastV \downtiny{[ECCV'24]} & 76.1 & 52.8 & 80.0 & 82.1 & 2130 & 76.1 & 92.9\% \\
    DART \downtiny{[EMNLP'25]} & 78.0 & 53.4 & 81.4 & 83.9 & \textbf{2245} & \textbf{78.9} & 95.5\% \\
    \rowcolor{gray!15}
    \textbf{V$^2$Drop} & \textbf{78.0} & \textbf{53.5} & \textbf{81.6} & \textbf{87.2} & 2224 & 78.7 & \textbf{96.0\%} \\
    \midrule
    \multicolumn{8}{c}{\textit{Token Reduction($\downarrow$77.8\%)}} \\
    \midrule
    FastV \downtiny{[ECCV'24]} & 73.8 & \textbf{49.3} & 78.3 & 79.2 & 2031 & 74.1 & 89.5\% \\
    DART \downtiny{[EMNLP'25]} & 74.4 & 48.5 & \textbf{79.6} & 82.1 & \textbf{2175} & 77.3 & 91.8\% \\
    \rowcolor{gray!15}
    \textbf{V$^2$Drop} & \textbf{75.6} & 48.7 & 78.9 & \textbf{85.1} & 2173 & \textbf{75.8} & \textbf{92.3\%} \\
    \bottomrule
    \end{tabular}
    \vspace{-2mm}
    }
    \captionof{table}{\textbf{Comparison with Qwen2-VL-7B across multiple image understanding benchmarks.} }
    \vspace{-3mm}
    \label{tab:main_results_qwen2}
\end{table}
\begin{table}[t!]

    \centering
    \centering
    \setlength\tabcolsep{3pt}
    \renewcommand{\arraystretch}{1.0}
    \vspace{0pt}
    {\fontsize{7.5pt}{7.5pt}\selectfont
    \definecolor{softgreen}{HTML}{66BB6A}   
    \definecolor{softred}{HTML}{E85C5C}     
    \definecolor{softgray}{HTML}{A6A6A6}    
    
    \newcommand{\downtiny}[1]{{\!\scriptsize{#1}}}
    \newcommand{\greentiny}[1]{\textcolor{softgreen}{#1}}
    \newcommand{\redtiny}[1]{\textcolor{softred}{#1}}
    \newcommand{\graytiny}[1]{\textcolor{softgray}{#1}}
    \begin{tabular}{lcccccc}
    \toprule
    \multirow{2}{*}{\textbf{Methods}} & \multirow{2}{*}{\textbf{MVBench}} & \multicolumn{4}{c}{\textbf{VideoMME}} & \multirow{2}{*}{\textbf{Avg.}} \\
    \cmidrule(lr){3-6}
     & & Overall & Short & Medium & Long & \\
    \midrule
    \multicolumn{7}{l}{\textit{Upper Bound, All Tokens (100\%)}} \\
    Qwen2-VL-7B~\cite{Wang:Qwen2-VL} & \graytiny{66.1} & \graytiny{57.7} & \graytiny{70.4} & \graytiny{54.6} & \graytiny{48.0} & \graytiny{100.0\%} \\
    \midrule
    \multicolumn{7}{c}{\textit{Average Retention Ratio = 20\%}} \\
    \midrule
    FastV \downtiny{[ECCV'24]} & 50.9 & 49.4 & 58.2 & 45.7 & 44.4 & 81.3\% \\
    DART \downtiny{[EMNLP'25]} & 58.9 & 53.0 & \textbf{64.1} & 49.4 & 45.4 & 90.5\% \\
    \rowcolor{gray!15}
    \textbf{V$^2$Drop} & \textbf{62.1} & \textbf{53.5} & 63.7 & \textbf{51.0} & \textbf{45.9} & \textbf{93.3\%} \\
    \bottomrule
    \end{tabular}
    \vspace{-2mm}
    }
    \captionof{table}{\textbf{Performance comparison with Qwen2-VL-7B across video understanding benchmarks.} 
    \vspace{-3mm}
    }
    \label{tab:video_results_2}
\end{table}

\begin{table}[t!]

    \centering
    \centering
    \setlength\tabcolsep{2.4pt}
    \renewcommand{\arraystretch}{1.0}
    \vspace{0pt}
    {\fontsize{7.5pt}{7.5pt}\selectfont
        
    \definecolor{softgreen}{HTML}{66BB6A}   
    \definecolor{softred}{HTML}{E85C5C}     
    \definecolor{softgray}{HTML}{A6A6A6}    
    
    \newcommand{\downtiny}[1]{{\!\scriptsize{#1}}}
    \newcommand{\greentiny}[1]{\textcolor{softgreen}{#1}}
    \newcommand{\redtiny}[1]{\textcolor{softred}{#1}}
    \newcommand{\graytiny}[1]{\textcolor{softgray}{#1}}

    \begin{tabular}{lcccccc}
    \toprule
    \multirow{2}{*}{\textbf{Methods}} & \multirow{2}{*}{\textbf{MVBench}} & \multicolumn{4}{c}{\textbf{VideoMME}} & \multirow{2}{*}{\textbf{Avg.}} \\
    \cmidrule(lr){3-6}
     & & Overall & Short & Medium & Long & \\
    \midrule
    \multicolumn{7}{l}{\textit{Upper Bound, All Tokens (100\%)}} \\
    LLaVA-OV-7B~\cite{li2024llava-ov} & \graytiny{56.9} & \graytiny{58.5} & \graytiny{70.0} & \graytiny{56.6} & \graytiny{48.9} & \graytiny{100.0\%} \\
    \midrule
    \multicolumn{7}{c}{\textit{Average Retention Ratio = 30\%}} \\
    \midrule
    DyCoke \downtiny{[CVPR'25]} & 56.6 & 56.1 & 67.1 & 54.6 & 46.6 & 97.7\% \\
    \midrule
    \multicolumn{7}{c}{\textit{Average Retention Ratio = 25\%}} \\
    \midrule
    FastV \downtiny{[ECCV'24]} & 55.5 & 55.3 & 65.0 & 53.8 & 47.0 & 96.0\% \\
    SparseVLM \downtiny{[ICML'25]} & 56.3 & 57.3 & 68.4 & 55.2 & 48.1 & 98.4\% \\
    PDrop \downtiny{[CVPR'25]} & 55.3 & 55.5 & 64.7 & 53.1 & 48.7 & 96.0\% \\
    DyCoke \downtiny{[CVPR'25]} & 49.5 & 51.0 & 61.1 & 48.6 & 43.2 & 87.1\% \\
    \rowcolor{gray!15}
    \textbf{V$^2$Drop} & \textbf{56.4} & \textbf{57.4} & \textbf{68.2} & \textbf{54.6} & \textbf{49.6} & \textbf{98.6\%} \\
    \midrule
    \multicolumn{7}{c}{\textit{Average Retention Ratio = 15\%}} \\
    \midrule
    FastV \downtiny{[ECCV'24]} & 51.6 & 48.1 & 51.4 & 49.4 & 43.3 & 86.5\% \\
    SparseVLM \downtiny{[ICML'25]} & 52.9 & 53.4 & 61.0 & 52.1 & 47.0 & 92.1\% \\
    PDrop \downtiny{[CVPR'25]} & 53.2 & 50.1 & 58.7 & 48.7 & 45.0 & 89.6\% \\
    \rowcolor{gray!15}
    \textbf{V$^2$Drop} & \textbf{53.9} & \textbf{54.4} & \textbf{64.1} & \textbf{51.4} & \textbf{47.7} & \textbf{93.9\%} \\
    \bottomrule
    \end{tabular}
    \vspace{-2mm}
    }
    \captionof{table}{\textbf{Performance comparison with LLaVA-OV-7B across video understanding benchmarks.} 
    \vspace{-3mm}
    }
    \label{tab:video_results}
\end{table}

\subsection{Main Comparisons}

\begin{table*}
    \vspace{-3mm}

    \centering
    \setlength\tabcolsep{4.2pt}  
    \renewcommand{\arraystretch}{1}
    {\fontsize{9pt}{9pt}\selectfont  

    \definecolor{softgreen}{HTML}{66BB6A}   
    \definecolor{softred}{HTML}{E85C5C}     
    \definecolor{softgray}{HTML}{A6A6A6}    
    
    \newcommand{\downtiny}[1]{{\!\scriptsize{#1}}}
    \newcommand{\greentiny}[1]{\textcolor{softgreen}{#1}}
    \newcommand{\redtiny}[1]{\textcolor{softred}{#1}}
    \newcommand{\graytiny}[1]{\textcolor{softgray}{#1}}
    \begin{tabular}{lcccccc}
    \toprule
    \multirow{2}{*}{\textbf{Methods}} & \textbf{LLM Generation↓} & \textbf{Model Generation↓} & \textbf{Total Latency↓} & \textbf{GPU Peak↓} & \textbf{Throughput↑} & \multirow{2}{*}{\textbf{Performance↑}} \\
     & \textbf{Latency (s)} & \textbf{Latency (s)} & \textbf{(min:sec)} & \textbf{Memory (MB)} & \textbf{(item/s)} &  \\
    \midrule
    \multicolumn{7}{l}{\textit{Upper Bound, 576 Tokens (100\%)}} \\
    LLaVA-1.5-7B~\cite{Liu:LLaVA-W} & \graytiny{400} & \graytiny{558} & \graytiny{10:14} & \graytiny{15566} & \graytiny{7.13} & \graytiny{64.6} \\
    \midrule
    Random & 270.4 \downtiny{\greentiny{(↓32.4\%)}} & 425.9 \downtiny{\greentiny{(↓23.7\%)}} & 8:02 \downtiny{\greentiny{(↓21.5\%)}} & 15045 \downtiny{\greentiny{(↓3.3\%)}} & 9.08 \downtiny{\greentiny{(↑1.27×)}} & 59.1 \downtiny{\greentiny{(91.5\%)}} \\
    \midrule
    FastV \downtiny{[ECCV'24]} & 294.0 \downtiny{\greentiny{(↓26.5\%)}} & 449.7 \downtiny{\greentiny{(↓19.4\%)}} & 8:26 \downtiny{\greentiny{(↓17.6\%)}} & 16139 \downtiny{\redtiny{(↑3.7\%)}} & 8.65 \downtiny{\greentiny{(↑1.21×)}} & 56.1 \downtiny{\greentiny{(86.8\%)}} \\
    SparseVLM \downtiny{[ICML'25]} & 288.2 \downtiny{\greentiny{(↓28.0\%)}} & 443.5 \downtiny{\greentiny{(↓20.5\%)}} & 8:20 \downtiny{\greentiny{(↓18.6\%)}} & 19229 \downtiny{\redtiny{(↑23.5\%)}} & 8.75 \downtiny{\greentiny{(↑1.23×)}} & 60.0 \downtiny{\greentiny{(92.9\%)}} \\
    PDrop \downtiny{[CVPR'25]} & 279.6 \downtiny{\greentiny{(↓30.1\%)}} & 429.7 \downtiny{\greentiny{(↓23.0\%)}} & 8:09 \downtiny{\greentiny{(↓20.3\%)}} & 15197 \downtiny{\greentiny{(↓2.3\%)}} & 8.95 \downtiny{\greentiny{(↑1.25×)}} & 61.1 \downtiny{\greentiny{(94.6\%)}} \\
    \rowcolor{gray!15}
    \textbf{V$^2$Drop} & \textbf{273.9} \textbf{\downtiny{\greentiny{(↓31.5\%)}}} & \textbf{429.3} \textbf{\downtiny{\greentiny{(↓23.1\%)}}} & \textbf{8:06} \textbf{\downtiny{\greentiny{(↓20.8\%)}}} & \textbf{15046} \textbf{\downtiny{\greentiny{(↓3.3\%)}}} & \textbf{9.01} \textbf{\downtiny{\greentiny{(↑1.26×)}}} & \textbf{61.8} \textbf{\downtiny{\greentiny{(95.7\%)}}} \\
    \toprule
    \multicolumn{7}{l}{\textit{Upper Bound, All Tokens (100\%)}} \\
    LLaVA-OV-7B~\cite{li2024llava-ov} & \graytiny{752.2} & \graytiny{1201.6} & \graytiny{32:02} & \graytiny{17686} & \graytiny{0.52} & \graytiny{56.9} \\
    \midrule
    Random & 190.9 \downtiny{\greentiny{(↓74.6\%)}} & 639.0 \downtiny{\greentiny{(↓46.8\%)}} & 23:09 \downtiny{\greentiny{(↓27.7\%)}} & 16298 \downtiny{\greentiny{(↓7.8\%)}} & 0.72 \downtiny{\greentiny{(↑1.38×)}} & 54.6 \downtiny{\greentiny{(96.0\%)}} \\
    \midrule
    FastV \downtiny{[ECCV'24]} & 315.9 \downtiny{\greentiny{(↓58.0\%)}} & 781.1 \downtiny{\greentiny{(↓35.0\%)}} & 25:05 \downtiny{\greentiny{(↓21.7\%)}} & 24619 \downtiny{\redtiny{(↑39.2\%)}} & 0.67 \downtiny{\greentiny{(↑1.29×)}} & 55.5 \downtiny{\greentiny{(97.5\%)}} \\
    SparseVLM \downtiny{[ICML'25]} & 493.8 \downtiny{\greentiny{(↓34.4\%)}} & 960.9 \downtiny{\greentiny{(↓20.0\%)}} & 30:12 \downtiny{\greentiny{(↓5.7\%)}} & 27378 \downtiny{\redtiny{(↑54.8\%)}} & 0.55 \downtiny{\greentiny{(↑1.06×)}} & 56.4 \downtiny{\greentiny{(99.1\%)}} \\
    PDrop \downtiny{[CVPR'25]} & 256.3 \downtiny{\greentiny{(↓65.9\%)}} & 719.0 \downtiny{\greentiny{(↓40.2\%)}} & 23:18 \downtiny{\greentiny{(↓27.3\%)}} & 24371 \downtiny{\redtiny{(↑37.8\%)}} & 0.71 \downtiny{\greentiny{(↑1.36×)}} & 55.3 \downtiny{\greentiny{(97.2\%)}} \\
    DyCoke \downtiny{[CVPR'25]} & 249.2 \downtiny{\greentiny{(↓66.7\%)}} & 713.5 \downtiny{\greentiny{(↓40.6\%)}} & 23:25 \downtiny{\greentiny{(↓26.9\%)}} & 16298 \downtiny{\greentiny{(↓7.8\%)}} & 0.71 \downtiny{\greentiny{(↑1.36×)}} & 49.5 \downtiny{\greentiny{(87.0\%)}} \\
    \rowcolor{gray!15}
    \textbf{V$^2$Drop} & \textbf{193.8} \textbf{\downtiny{\greentiny{(↓74.2\%)}}} & \textbf{642.4} \textbf{\downtiny{\greentiny{(↓46.5\%)}}} & \textbf{23:13} \textbf{\downtiny{\greentiny{(↓27.5\%)}}} & \textbf{16298} \textbf{\downtiny{\greentiny{(↓7.8\%)}}} & \textbf{0.72} \textbf{\downtiny{\greentiny{(↑1.38×)}}} & \textbf{56.4} \textbf{\downtiny{\greentiny{(99.1\%)}}} \\
    \bottomrule
    \end{tabular}
    }
    \vspace{-2mm}
    \captionof{table}{\textbf{Efficiency comparison on image/video understanding.} We measure: (1) LLM Generation Latency: LLM-only response time; (2) Model Generation Latency: full model response time; (3) Total Latency: time to complete MMBench/MVBench on LLaVA-1.5-7B/LLaVA-OV-7B; (4) Throughput: samples processed per second.}
    \vspace{-2mm}
    \label{tab:efficiency_comparison_merged}
\end{table*}

\vspace{0.3em}
\noindent \textbf{Image Understanding.}  Table~\ref{tab:image_comparison_results_1} compares V$^2$Drop with existing methods across multiple benchmarks using LLaVA-1.5-7B at different retention ratios. The upper section of Table~\ref{tab:efficiency_comparison_merged} presents the inference efficiency of V$^2$Drop on LLaVA-1.5-7B. Considering both performance and efficiency, our analysis reveals three key findings: \textbf{(i) State-of-the-art Performance:} With only 192 tokens retained (66.7\% reduction), V$^2$Drop achieves an impressive average performance of \textbf{97.6\%}, substantially outperforming the second-best method PDrop by \textbf{1.6\%}. Even under more aggressive reduction ratios, V$^2$Drop maintains competitive performance. \textbf{(ii) Efficient Operator Compatibility:} V$^2$Drop eliminates explicit attention score computation, enabling seamless integration with Flash Attention~\cite{daoFlashAttention-2}. Without introducing additional memory overhead, V$^2$Drop achieves peak memory usage and total latency comparable to random token dropping. \textbf{(iii) Seamless Scalability to Advanced Models:} As shown in Table~\ref{tab:main_results_qwen2}, V$^2$Drop consistently outperforms FastV and DART across nearly all benchmarks on Qwen2-VL~\cite{Wang:Qwen2-VL} under different configurations, demonstrating effectiveness at high resolutions and compatibility with variable-resolution inputs.

\begin{figure*}[t]
    \centering
    \includegraphics[width=\textwidth]{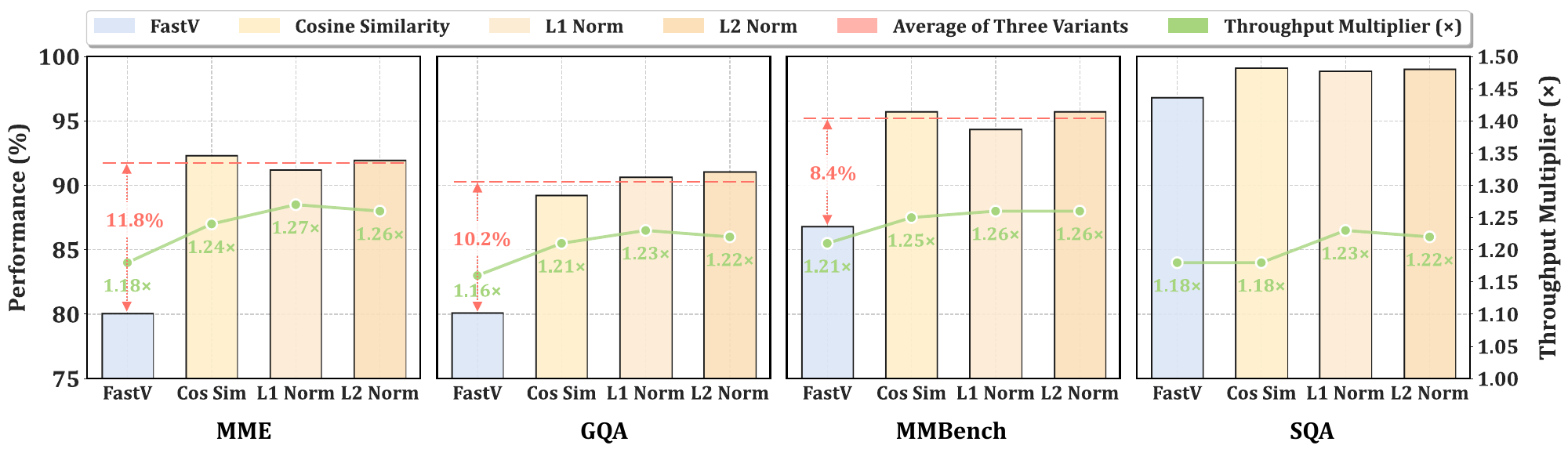}
    \vspace{-7mm}
    \captionof{figure}{\textbf{Effects of different variation measurement metrics.} Comparison of three variation measurement methods with FastV when retaining 128 tokens on LLaVA-1.5-7B across different datasets. The red line represents the average performance gap between the three strategies and FastV, while the green line shows throughput. 
    }
    \vspace{-4mm}
    \label{fig:ablation1}
\end{figure*}

\vspace{0.5em}
\noindent \textbf{Video Understanding.} We further extend V$^2$Drop to video understanding using LLaVA-OV-7B and Qwen2-VL-7B. Table~\ref{tab:video_results_2} and Table~\ref{tab:video_results} compare V$^2$Drop with state-of-the-art token compression methods across multiple benchmarks, while the lower section of Table~\ref{tab:efficiency_comparison_merged} reports inference efficiency. Our analysis reveals three key findings: \textbf{(i) Superior Performance:} V$^2$Drop outperforms all competing methods across all benchmarks on LLaVA-OV and Qwen2-VL, achieving \textbf{98.6\%} of original performance with only 25\% token retention, surpassing DyCoke (97.7\% with 30\% tokens). At aggressive compression ($R=15\%$), V$^2$Drop maintains exceptional robustness while baseline methods degrade significantly. \textbf{(ii) Excellence in Long Video Understanding:} V$^2$Drop significantly outperforms baselines on long video tasks such as VideoMME (Long) by mitigating positional bias problem, where VideoLLMs disproportionately focus on later-frame tokens. \textbf{(iii) Superior Inference Efficiency:} V$^2$Drop maintains high throughput while reducing GPU peak memory.
In contrast, we surprisingly find that SparseVLM increases peak memory by \textbf{54.8\%} on MVBench due to its merging strategy and explicit attention computation, greatly elevating computational costs, while our V$^2$Drop inherently avoids such operations.

\subsection{Ablation Study}
\label{subsec:ablation}

\vspace{0.3em}
\noindent \textbf{Effects of Variation Metric.} Figure~\ref{fig:ablation1} presents a analysis of three variation metrics against FastV across multiple benchmarks. All three metrics---Cosine Similarity, L1 Norm, and L2 Norm---outperform FastV in average performance, validating the robustness of variation-based dropping strategies. Furthermore, all metrics demonstrate notable improvements in inference throughput, with L2 Norm achieving the optimal balance between performance and efficiency.

\begin{figure*}[!t]
    \centering
    \includegraphics[width=\textwidth]{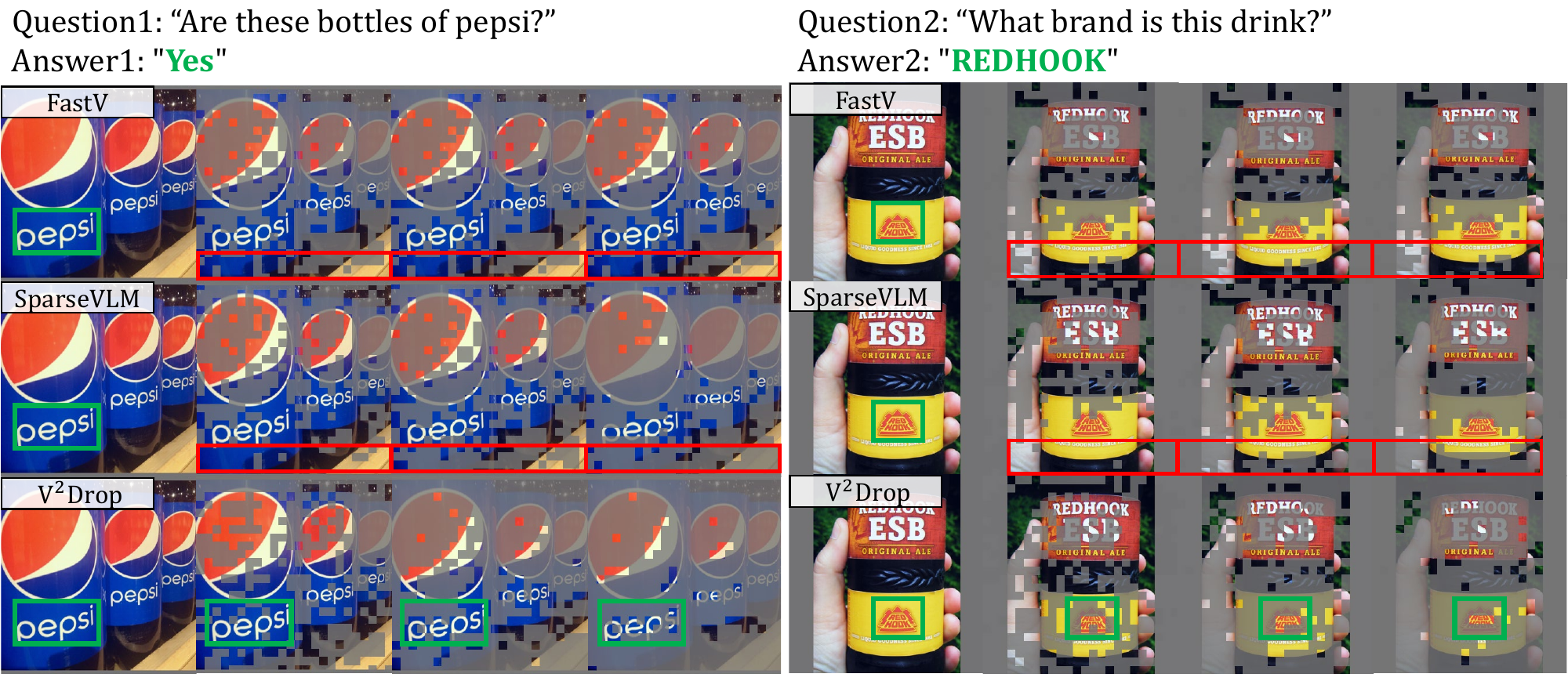}
    \vspace{-7mm}
    \captionof{figure}{\textbf{Visualization of token compression.} Rows 1, Rows 2 and Rows 3 present compressed results from FastV, SparseVLM and our V$^2$Drop respectively, where grey masks indicate discarded tokens.}
    \vspace{-5mm}
    \label{fig:visualization}
\end{figure*}

\begin{figure}[!t]
    \centering
    \includegraphics[width=\linewidth]
    {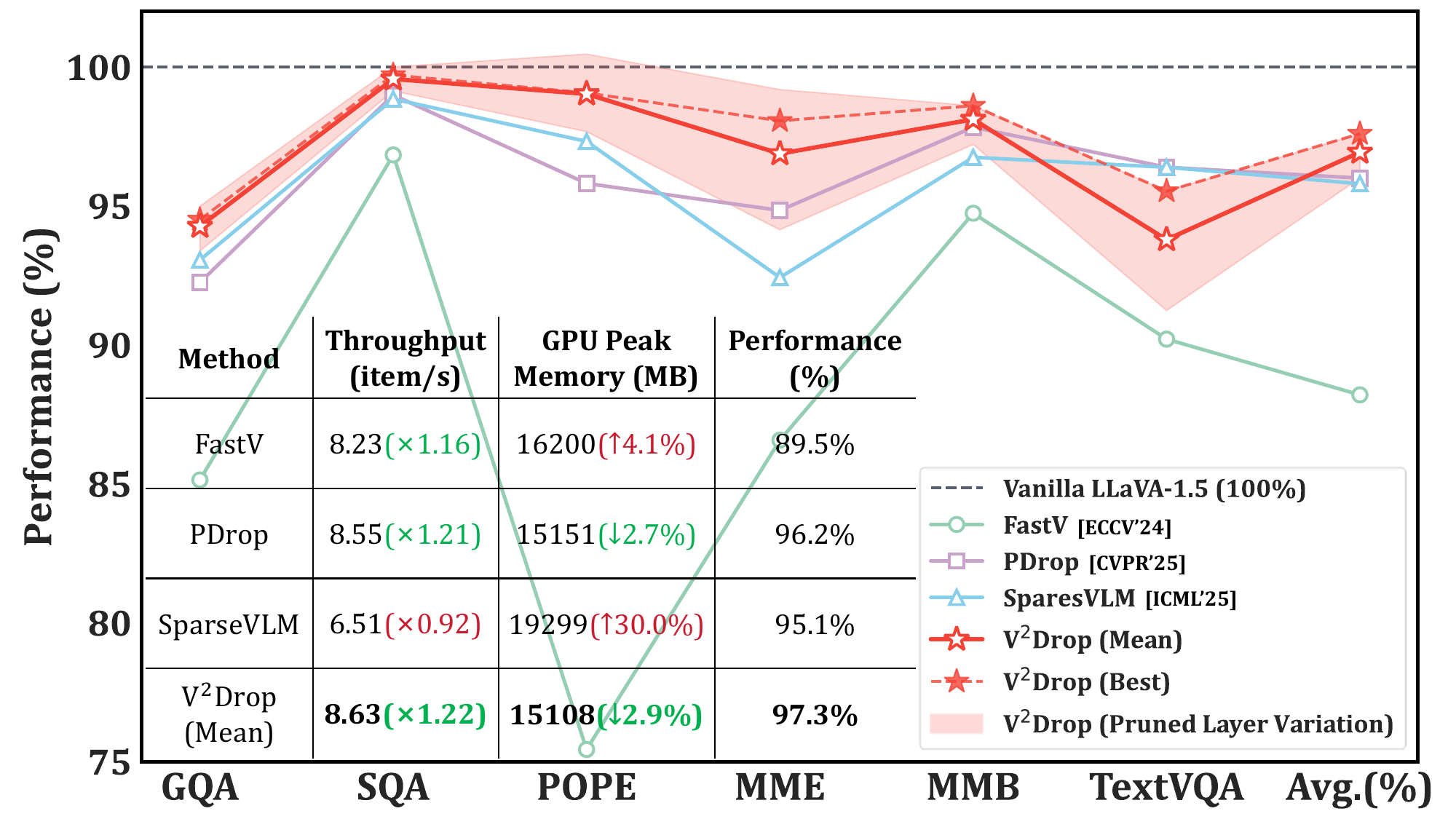}
    \vspace{-4mm}
    \captionof{figure}{\textbf{Effects of token dropping across layers.} Performance with 192 retained tokens on LLaVA-1.5-7B across datasets, and efficiency analysis on MMB.}
    \vspace{-4mm}
    \label{fig:ablation2}
\end{figure}

\vspace{0.3em}
\noindent \textbf{Effects of Token Dropping across Different Layers.} We investigate the effects of token dropping across different LLM transformer layers. As shown in Figure~\ref{fig:ablation2}, V$^2$Drop achieves \textbf{97.3\%} of vanilla performance on average across all benchmarks and consistently outperforms FastV and SparseVLM under varying dropping layers. Furthermore, V$^2$Drop exhibits superior inference efficiency, delivering substantial throughput improvements and GPU peak memory reduction while maintaining competitive performance.

\begin{figure}[!t]
    \centering
    \includegraphics[width=\linewidth]{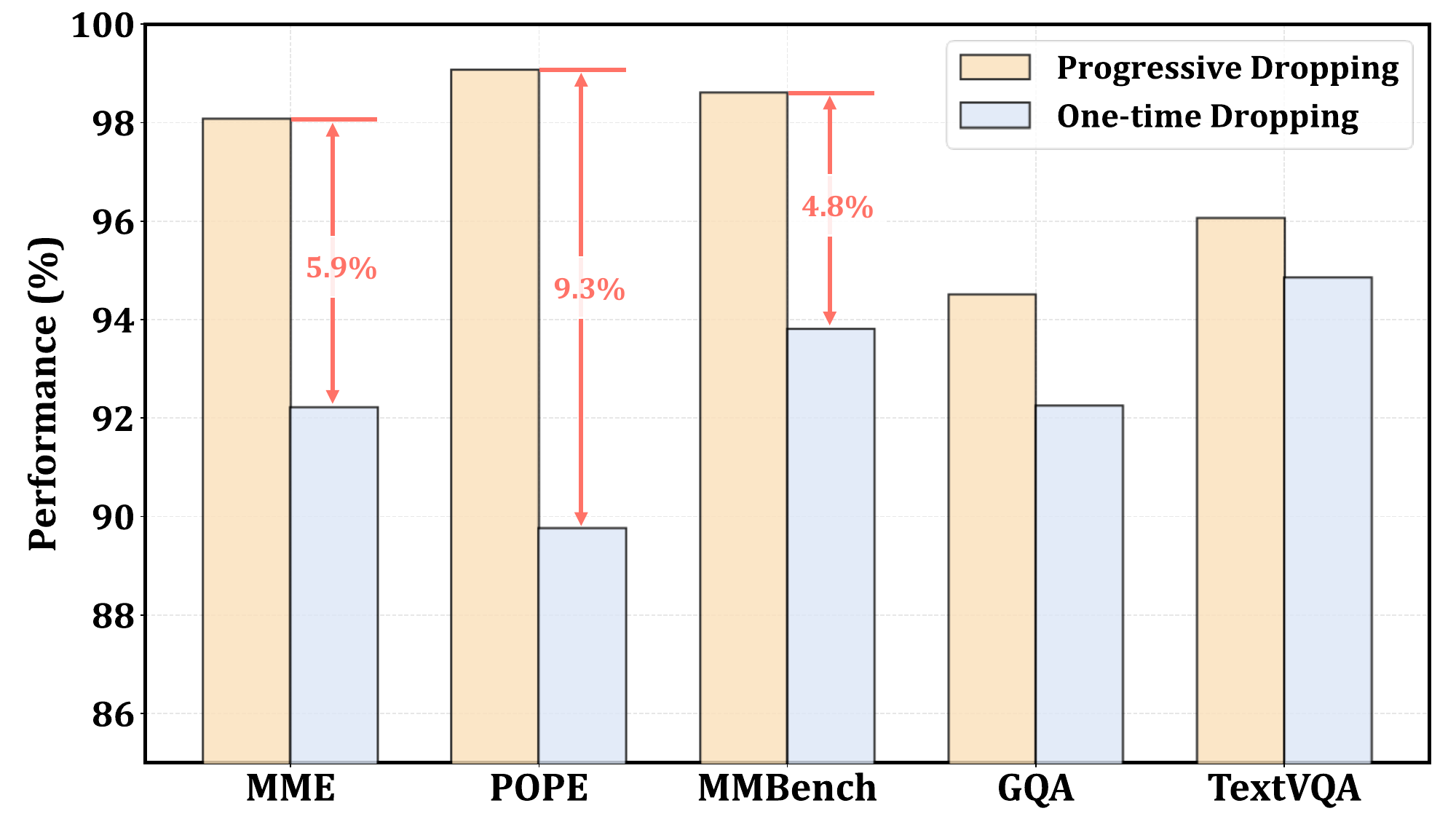}
    \vspace{-6mm}
    \captionof{figure}{\textbf{Effects of progressive token dropping.} Performance with 192 retained tokens on LLaVA-1.5-7B.}
    \vspace{-4mm}
\label{fig:ablation3}
\end{figure}

\vspace{0.3em}
\noindent \textbf{Effects of Progressive Token Dropping.} We investigate the impact of two token pruning strategies on V$^2$Drop: progressive dropping and one-time dropping. As shown in Figure~\ref{fig:ablation3}, progressive dropping outperforms one-time dropping, achieving gains of \textbf{5.9\%} and \textbf{9.3\%} on MME and POPE, respectively. These results validate that progressive dropping more effectively preserves visual information through gradual selection, mitigating the risk of prematurely discarding important features while reducing computational overhead.

\subsection{Visualizations}

Figure~\ref{fig:visualization} illustrates token compression effects on TextVQA. From left to right, we visualize results after pruning at different layers. FastV employs one-time dropping, retaining redundant tokens in subsequent layers. SparseVLM adopts progressive dropping but exhibits positional bias, preferentially retaining later-sequence tokens and causing information loss. In contrast, V$^2$Drop preserves critical tokens based on variation, progressively selecting semantically important regions while avoiding positional bias.

\section{Conclusion}
\label{sec:Conclusion}

Large vision-language models (LVLMs) require substantial computational resources due to extensive visual tokens during inference. Existing compression methods often rely on attention weights or introduce positional bias, limiting their effectiveness and compatibility with efficient operators. We present V$^2$Drop, a novel method addressing these limitations through a token variation perspective. We reveal that visual token variations exhibit task-agnostic properties, enabling compression without attention weights or positional bias. V$^2$Drop progressively removes minimal-variation tokens, maintaining compatibility with efficient operators while achieving significant computational savings. Experiments demonstrate V$^2$Drop effectively balances performance and efficiency across benchmarks.
{
    \small
    \bibliographystyle{ieeenat_fullname}
    \bibliography{main}
}

\appendix
\clearpage
\setcounter{page}{1}
\maketitlesupplementary

In the appendix, we provide detailed experimental settings in Section~\ref{sec:appendix/Experimental_Settings}, additional experimental results in Section~\ref{sec:appendix/Additional_Experimental_Results}, algorithmic descriptions in Section~\ref{sec:appendix/Algorithm}, further discussion on content-agnostic positional bias in Section~\ref{sec:appendix/more_dis}, and detailed theoretical analysis in Section~\ref{sec:appendix/Theoretical_Analysis}.

\section{Detailed Experimental Settings}
\label{sec:appendix/Experimental_Settings}
\vspace{0.5em}
\noindent \textbf{Benchmark Details.} We evaluate V$^2$Drop on various multi-modal understanding benchmarks detailed as follows:

\begin{itemize}

    \item \textbf{GQA}~\cite{Hudson:GQA} comprises scene graphs, questions, and images, designed to test visual scene understanding and multi-aspect image reasoning capabilities.

    \item \textbf{MMBench}~\cite{Liu:MMBench} evaluates models through a three-level hierarchical structure with 20 specific ability dimensions, enabling comprehensive assessment of perception and reasoning capabilities.

    \item \textbf{MME}~\cite{Fu:MME} comprises 14 subtasks evaluating perceptual and cognitive abilities through manually constructed instruction-answer pairs, mitigating data leakage issues.

    \item \textbf{POPE}~\cite{Li:POPE} evaluates object hallucination through binary questions about object presence, using accuracy, recall, precision, and F1 metrics across three sampling strategies.

    \item \textbf{ScienceQA}~\cite{Lu:ScienceQA} spans natural, language, and social sciences with hierarchical categorization, evaluating multimodal understanding and multi-step reasoning capabilities.

    \item \textbf{TextVQA}~\cite{Singh:TextVQA} evaluates models' ability to read and reason about text within images through visual question-answering tasks requiring integrated textual understanding.

    \item \textbf{AI2D}~\cite{kembhavi2016AI2D}comprises 5,000 scientific diagrams with accompanying questions that test visual-spatial reasoning capabilities across educational content.

    \item \textbf{MMStar}~\cite{chen2024:MMStar}provides 12,000 high-resolution images designed to evaluate spatial, temporal, and commonsense reasoning across multimodal understanding tasks.

    \item \textbf{MVBench}~\cite{li2024mvbench} defines 20 video understanding tasks that require deep comprehension of temporal dimensions, beyond single-frame analysis.

    \item \textbf{VideoMME}~\cite{fu2024videomme} comprises 900 videos and 2,700 multiple-choice questions across six domains, with durations from 11 seconds to 1 hour, categorized into short, medium, and long subsets.

\end{itemize}                                                                                                                                                     
\vspace{3em}
\noindent \textbf{Baseline Models.} The baseline LVLMs, as follows:

\begin{itemize}

    \item \textbf{LLaVA-1.5}~\cite{Liu:LLaVA-1.5} enhances multimodal understanding by scaling visual instruction tuning with academic-task-oriented datasets and improved training recipes. It incorporates a two-stage training approach that first aligns vision and language representations, then fine-tunes on diverse instruction-following data, achieving strong performance on visual reasoning, OCR, and multimodal dialogue tasks across various benchmarks.
        
    \item \textbf{Qwen2-VL}~\cite{Wang:Qwen2-VL} enhances multimodal perception by investigating scaling laws for vision-language models. By scaling model size (2B, 8B, and 72B parameters) and training data, it achieves competitive performance across diverse tasks. It supports any resolution input, enabling superior performance on document parsing, OCR, visual reasoning, and video understanding while maintaining strong text-image alignment.

    \item \textbf{LLaVA-OneVision}~\cite{li2024llava-ov} unifies single-image, multi-image, and video tasks in a single model. It represents videos as long visual token sequences in the same ``interleaved'' format used for images, enabling smooth task transfer from images to videos and facilitating strong zero-shot video understanding capabilities.

\end{itemize}
\vspace{1em}
\noindent \textbf{Comparison Methods.} We provide detailed introductions and comparisons of existing token compression methods mentioned in the main text, as follows:

\begin{table*}[!t]
    \centering
    \setlength\tabcolsep{6.2pt}
    \renewcommand{\arraystretch}{1.0}
    {\fontsize{9pt}{9pt}\selectfont
    \begin{tabular}{@{\extracolsep{\fill}}c|c|cccccc|ccc} 
    \hline
    \rule{0pt}{2.6ex}  
    \multirow{2}{*}{\textbf{Benchmark}} & \multirow{2}{*}{\textbf{Vanilla}} 
        & \multicolumn{6}{c|}{\textbf{Pruned Layers Selection}} 
        & \multicolumn{3}{c}{\textbf{Other Methods}} \\
    \cline{3-11}
    \rule{0pt}{2.6ex}  
    &
        & (4,14,30) & (3,14,29) & (3,15,27) & (3,16,24) & (3,17,22) & (2,16,21) 
        & FastV & SparseVLM & PDrop \\
    \hline
    \rule{0pt}{2.6ex}  
    GQA & 61.9 & 57.8 & 58.6 & 58.5 & 58.8 & 58.5 & 57.9 & 52.7 & 57.1 & 57.6 \\
    SQA & 69.5 & 68.9 & 69.1 & 69.3 & 69.1 & 69.3 & 69.5 & 67.3 & 68.8 & 68.7 \\
    POPE & 85.9 & 86.3 & 85.0 & 85.1 & 85.0 & 85.1 & 83.9 & 64.8 & 82.3 & 83.6 \\
    MME & 1862 & 1753 & 1826 & 1847 & 1813 & 1826 & 1759 & 1612 & 1766 & 1721 \\
    MMB & 64.6 & 63.2 & 63.4 & 63.7 & 63.5 & 63.7 & 62.8 & 61.2 & 63.2 & 62.5 \\
    TextVQA & 58.2 & 53.1 & 54.0 & 54.8 & 55.2 & 55.6 & 54.8 & 52.5 & 56.1 & 56.1 \\
    \hline
    \rule{0pt}{2.5ex} Avg. (\%) & 100.0\% & 96.0\% & 97.0\% & 97.5\% & 97.3\% & 97.6\% & 96.2\% & 88.2\% & 96.0\% & 95.8\% \\
    [0.4ex]
    \hline
    \end{tabular}
    }
    \vspace{-1.5mm}
    \caption{\textbf{Supplementary results on pruned layers selection.} Performance with 192 retained tokens on LLaVA-1.5-7B across datasets. The notation (a, b, c) represents a three-stage pruning strategy with token reduction applied at the a-th, b-th, and c-th layers, respectively.}
    \vspace{-2.8mm}
    \label{tab:ablation2}
\end{table*}

\begin{table}[h]
\centering
\fontsize{9pt}{9pt}\selectfont 
\renewcommand{\arraystretch}{1.2}

\definecolor{softgreen}{HTML}{66BB6A}   
\definecolor{softred}{HTML}{E85C5C}     
\definecolor{softgray}{HTML}{A6A6A6}    
    
\newcommand{\downtiny}[1]{{\!\scriptsize{#1}}}
\newcommand{\greentiny}[1]{\textcolor{softgreen}{#1}}
\newcommand{\redtiny}[1]{\textcolor{softred}{#1}}
\newcommand{\graytiny}[1]{\textcolor{softgray}{#1}}

\setlength{\tabcolsep}{1.8pt}
\begin{tabular}{lcccc}
\hline
\multirow{2}{*}{Methods} & \multicolumn{4}{c}{Throughput (item/s)} \\
\cline{2-5}
& MME & GQA & MMBench & SQA \\
\hline
LLaVA-1.5-7B & 8.02 & 7.5 & 7.13 & 6.9 \\
\hline
FastV & 9.46 \downtiny{\greentiny{(1.18×)}} & 8.68 \downtiny{\greentiny{(1.16×)}} & 8.65 \downtiny{\greentiny{(1.21×)}} & 8.14 \downtiny{\greentiny{(1.18×)}} \\
Cosine Similarity & 9.95 \downtiny{\greentiny{(1.24×)}} & 9.13 \downtiny{\greentiny{(1.21×)}} & 8.90 \downtiny{\greentiny{(1.25×)}} & 8.14 \downtiny{\greentiny{(1.18×)}} \\
L1 Norm & 10.16 \downtiny{\greentiny{(1.27×)}} & 9.23 \downtiny{\greentiny{(1.23×)}} & 9.01 \downtiny{\greentiny{(1.26×)}} & 8.49 \downtiny{\greentiny{(1.23×)}} \\
\rowcolor{gray!15}
L2 Norm & 10.11 \downtiny{\greentiny{(1.26×)}} & 9.18 \downtiny{\greentiny{(1.22×)}} & 9.01 \downtiny{\greentiny{(1.26×)}} & 8.42 \downtiny{\greentiny{(1.22×)}} \\
\hline
\end{tabular}
\caption{\textbf{Supplementary results on variation metric selection. }Throughput with 128 retained tokens on LLaVA-1.5-7B across datasets. The notation (N$\times$) represents an N-fold throughput improvement compared to the baseline model LLaVA-1.5-7B.}
\vspace{-7.5mm}
\label{tab:ablation1-1}
\end{table}
\begin{table}[h]
\vspace{7mm}
\centering
\small
\renewcommand{\arraystretch}{1.2}
\newcommand{\downtiny}[1]{{\!\scriptsize{#1}}}
\setlength{\tabcolsep}{2pt}
\begin{tabular}{lccccc}
\hline
\multirow{2}{*}{Methods} & \multicolumn{5}{c}{Performance} \\
\cline{2-6}
& MME & POPE & MMBench & GQA & TextVQA \\
\hline
LLaVA-1.5-7B & 1862 & 85.9 & 64.6 & 61.9 & 58.2 \\
\hline
One-time dropping & 1717 & 77.1 & 60.6 & 57.1 & 55.2 \\
\rowcolor{gray!15}
Progressive dropping & \textbf{1826} & \textbf{85.1} & \textbf{63.7} & \textbf{58.5} & \textbf{55.9} \\
\hline
\end{tabular}
\caption{\textbf{Supplementary results on effects of
progressive token dropping. }Performance with 192 retained tokens on LLaVA-1.5-7B across datasets.}
\label{tab:ablation3}
\end{table}

\begin{itemize}

    \item \textbf{ToMe}~\cite{Bolya:ToMe} merges similar tokens in visual transformer layers through lightweight matching techniques, achieving acceleration without requiring additional training.

    \item \textbf{LLaVA-PruMerge}~\cite{Shang:LLaVA-PruMerge} combines pruning and merging strategies by dynamically removing less important tokens using CLS-patch attention and clustering retained tokens based on key similarity.

    \item \textbf{FastV}~\cite{Chen:FastV} focuses on early-stage token pruning by leveraging attention maps, effectively reducing computational overhead in the initial layers.

    \item \textbf{DART}~\cite{wen2025dart} introduces a duplication-aware token pruning approach that selects tokens based on their redundancy relative to pivot tokens rather than importance scores.

    \item \textbf{HiRED}~\cite{arif2024hired} allocates token budgets across image partitions based on CLS token attention, followed by the selection of the most informative tokens within each partition, ensuring spatially aware token reduction.

    \item \textbf{PDrop}~\cite{xing2024pdrop} adopts a progressive token-dropping strategy across model stages, forming a pyramid-like token structure that balances efficiency and performance.
    
    \item \textbf{SparseVLM}~\cite{Zhang:SparseVLM} ranks token importance using cross-modal attention and introduces adaptive sparsity ratios, complemented by a novel token recycling mechanism.
    Based on the abstract, here's a one-sentence summary without numbers:
    
    \item \textbf{DyCoke}~\cite{tao2024dycoke} is a two-stage VideoLLM method that prunes similar tokens temporally and compresses less-attended visual tokens in KV cache using LLM attention weights. Its reliance on frame-set division and similarity-based compression limits aggressive token compression, and while compatible with Flash Attention~\cite{daoFlashAttention-2}, it requires explicit attention weights making it incompatible with efficient attention operators.

\end{itemize}

\begin{table}[h]
\centering
\fontsize{9pt}{9pt}\selectfont 
\renewcommand{\arraystretch}{1.2}
\newcommand{\downtiny}[1]{{\!\scriptsize{#1}}}
\setlength{\tabcolsep}{7.5pt}
\begin{tabular}{lcccc}
\hline
\multirow{2}{*}{Methods} & \multicolumn{4}{c}{Performance} \\
\cline{2-5}
& MME & GQA & MMBench & SQA \\
\hline
LLaVA-1.5-7B & 1862 & 61.9 & 64.6 & 69.5 \\
\hline
FastV & 1490 & 49.6 & 56.1 & 67.3 \\
Cosine Similarity & 1718 & 55.2 & 61.8 & 69.2 \\
L1 Norm & 1698 & 56.1 & 60.9 & 68.7 \\
\rowcolor{gray!15}
L2 Norm & 1712 & 56.3 & 61.8 & 68.8 \\
\hline
\end{tabular}
\caption{\textbf{Supplementary results on variation metric selection. }Performance with 128 retained tokens on LLaVA-1.5-7B across datasets. }
\vspace{-4mm}
\label{tab:ablation1-2}
\end{table}

\vspace{1em}
\noindent \textbf{Implementation Details.} Our experiments are conducted on NVIDIA A100-PCIe-80GB GPUs. The implementation was carried out in Python 3.10, utilizing PyTorch 2.1.2 and CUDA 12.1. All baseline settings follow the original paper.

\vspace{1em}
\noindent \textbf{Experimental parameter details for V$^2$Drop.} On LLaVA-1.5-7B, we conduct three-stage pruning at layers 3, 17, and 22. When retaining 192 tokens, we prune 50\%, 70\%, and 100\% of Vision tokens at layers 3, 17, and 22. When retaining 128 tokens, we prune 72\%, 75\%, and 100\% of Vision tokens at layers 3, 17, and 22, respectively. When retaining 64 tokens, we prune 95\%, 95\%, and 100\% of Vision tokens at layers 3, 17, and 22, respectively.

\section{Additional Experimental Results}
\label{sec:appendix/Additional_Experimental_Results}

\noindent \textbf{Supplementary Results on Variation Metric Selection.} This section presents detailed results of V$^2$Drop from the ablation study on the Effects of Variation Metric. Table~\ref{tab:ablation1-1} and Table~\ref{tab:ablation1-2} respectively report the throughput and performance data of three variation metrics across multiple datasets. The experiments demonstrate that variation-based pruning strategies outperform attention-score-based pruning methods such as FastV in both performance and efficiency, validating the robustness of variation-based dropping strategies. For more intuitive visualizations, please refer to the main discussion in \S4.3.

\vspace{1em}
\noindent \textbf{Supplementary Results on Pruned Layers Selection.} This section presents comprehensive experimental results on LLaVA-1.5-7B, providing a detailed analysis of the pruning layer selection strategies of V$^2$Drop. Table~\ref{tab:ablation2} comprehensively lists performance metrics across multiple benchmarks, including GQA, SQA, POPE, MME, MMB, and TextVQA, with all experiments retaining 192 visual tokens. These findings further validate the robustness of V$^2$Drop under various pruning layer combinations. The table also includes comparisons with baseline methods such as FastV, SparseVLM, and PDrop, highlighting the consistent superiority of V$^2$Drop across different configurations. For more intuitive visualizations, please refer to the main discussion in §4.3.

\vspace{1em}
\noindent \textbf{Supplementary Results on Effects of Progressive Token Dropping.} This section presents detailed results of V$^2$Drop from the ablation study on the Effects of Progressive Token Dropping. Table~\ref{tab:ablation3} shows the performance of two token pruning strategies in V$^2$Drop: progressive token dropping and one-time dropping, evaluated on LLaVA-1.5-7B. The experiments demonstrate that progressive token dropping significantly outperforms one-time dropping across all datasets, proving that progressive token dropping more effectively preserves critical visual information through its gradual selection mechanism. For more intuitive visualizations, please refer to the main discussion in \S4.3.

\begin{figure*}[!t]
    \centering
    \includegraphics[width=\textwidth]{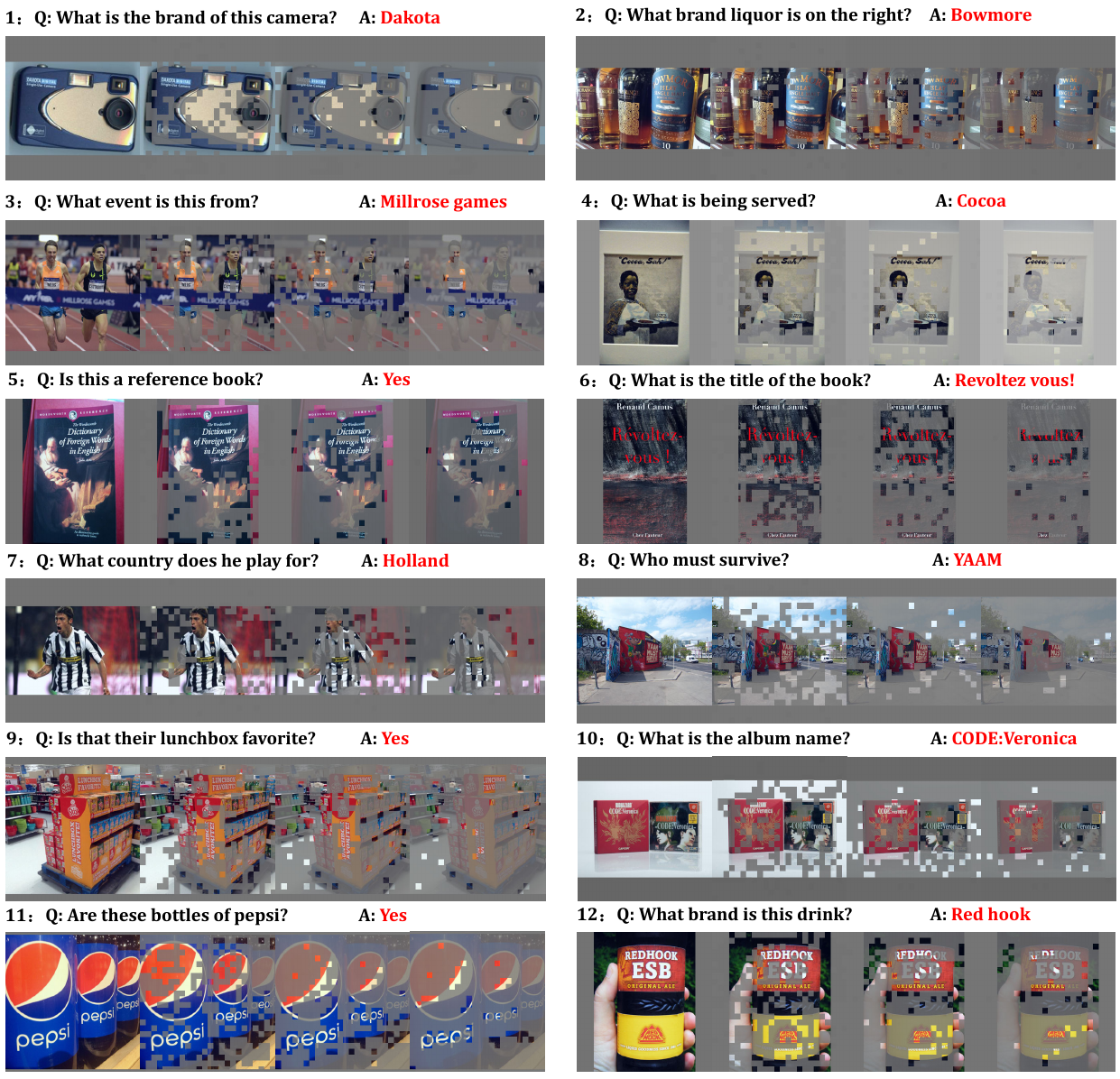}
    
    \captionof{figure}{\textbf{More visualization of token compression by V$^2$Drop.} The presented examples are from TextVQA, where grey masks indicate discarded visual tokens.}

    \label{fig:more visualization}
\end{figure*}

\vspace{1em}
\noindent \textbf{More Visualizations of Token Compression.} In Figure~\ref{fig:more visualization}, we present additional token compression visualization results of V$^2$Drop across diverse scenarios. The visualizations demonstrate that by preserving key tokens based on visual token variation information, V$^2$Drop progressively selects core tokens from images and focuses on semantically critical regions. This indicates that our token variation metric can effectively localize important regions. As illustrated in the figure, across various real-world scenarios where critical regions are located at different positions within the image—such as the bottom-left (case 11), top-right (case 5), and top-left (case 1)—our method consistently establishes accurate correspondence between token importance and semantic relevance.

\section{Algorithm Details of V$^2$Drop}
\label{sec:appendix/Algorithm}

Algorithm~\ref{alg:v2drop} presents the algorithm workflow of our V$^2$Drop method. This algorithm details the step-by-step process of our token compression approach, illustrating how V$^2$Drop dynamically compresses visual tokens based on variation analysis.

\begin{algorithm}[t]
\caption{V$^2$Drop: Variation-aware Vision Token Dropping}
\label{alg:v2drop}
\begin{algorithmic}[1]
\Require Vision tokens $\mathbf{F}^v \in \mathbb{R}^{M \times D'}$, Dropping layers $\mathcal{L} = \{l_1, l_2, \ldots, l_K\}$, Compression Targets $\{K_{l_1}, K_{l_2}, \ldots, K_{l_K}\}$
\Ensure Compressed vision tokens
\State Current token count $M_{\text{curr}} \leftarrow M$
\For{$l = 1, 2, \ldots, L$}
\If{$l \in \mathcal{L}$}
\vspace{0.3em}
\State \textbf{Step 1: Variation Computation}
\vspace{0.3em}
\For{$i = 1$ to $M_{\text{curr}}$}
\State $s_i^{(l)} \leftarrow \|\mathbf{f}_i^{(l)} - \mathbf{f}_i^{(l-1)}\|_2$
\EndFor
\State $\mathbf{S}^{(l)} = \{s_1^{(l)}, s_2^{(l)}, \ldots, s_{M_{\text{curr}}}^{(l)}\}$
\vspace{0.3em}
\State \textbf{Step 2: Token Ranking and Selection}
\vspace{0.3em}
\State $\text{indices} \leftarrow \mathrm{argsort}(\mathbf{S}^{(l)}, \text{descending})$
\State $\hat{\mathbf{F}}^v_l \leftarrow \{\mathbf{f}_{\text{indices}[j]}^{(l)} : j = 1, \ldots, K_l\}$
\State $\mathbf{F}^v_{\text{curr}} \leftarrow \hat{\mathbf{F}}^v_l$, $M_{\text{curr}} \leftarrow K_l$
\Else
\State $\mathbf{F}^v_{\text{curr}} \leftarrow \mathrm{TransformerLayer}(\mathbf{F}^v_{\text{curr}})$
\EndIf
\EndFor
\State \Return $\mathbf{F}^v_{\text{curr}}$
\end{algorithmic}
\end{algorithm}

\section{More Discussions about Content-agnostic
Positional Bias.}
\label{sec:appendix/more_dis}

Figures~\ref{fig:position_bias} and~\ref{fig:position_bias_quantity} reveal the inherent content-agnostic positional bias of LLM attention-guided methods such as SparseVLM and FastV. Figure~\ref{fig:position_bias} illustrates how these methods, despite assigning higher scores to critical regions, disproportionately favor later-positioned tokens regardless of content relevance, leading to the discarding of informative earlier tokens and triggering multimodal hallucinations. In contrast, measuring token-wise variation (\textit{e.g.}, L2 Norm) intuitively reflects token importance and selectively retains semantically critical tokens. To quantify this bias, Figure~\ref{fig:position_bias_quantity} analyzes LLaVA-1.5-7B and Qwen2-VL-7B across three datasets (TextVQA, POPE, and MME), partitioning tokens into 10 equal intervals and calculating retention probabilities after pruning 50\% of tokens at the third layer. Results demonstrate that attention-guided methods exhibit strong end-of-sequence bias, while variation-aware evaluation produces naturally uniform spatial distributions. Below, we provide a detailed theoretical analysis to establish the relationship between token variation and model output.

\section{Suppleymentary Theoretical Analysis}
\label{sec:appendix/Theoretical_Analysis}

Here, we present the complete theoretical proof that rigorously establishes the connection between token variation and model output through first-order analysis.

\subsection{Smoothness Assumption}
We assume the model $f$ has sufficient 
local smoothness in the representation space, such that the second-order 
remainder term in the Taylor expansion is bounded, satisfies: 
\begin{align}
\|R_j\| = \mathcal{O}(\|\Delta x_j^{(t)}\|^2).
\end{align}
This assumption is well-
justified in Transformer-based LVLMs due to three architectural properties:

\begin{itemize}[leftmargin=*,nosep]

    \item \textbf{Residual connections} limit layer-wise changes, ensuring 
    $\|\Delta x_j^{(t)}\|$ remains small relative to $\|x_j^{(t)}\|$;
    
    \item \textbf{Layer normalization} constrains the range of token 
    
    representations, bounding higher-order derivatives;
    \item \textbf{Smooth activations} (e.g., GELU, SiLU) provide continuous 
    second derivatives, ensuring Taylor expansion validity.
    
\end{itemize}

Under this assumption, 
for sufficiently small $\|\Delta x_j^{(t)}\|$, the quadratic term is negligible 
compared to the linear term, yielding:
\begin{align}
\|\Delta f_j\| \approx \|J_j\|_{\text{op}} \cdot \|\Delta x_j^{(t)}\|
\end{align}

\subsection{Justification of Bounded Jacobian Assumption}

In the proof of Corollary, we assume that for all tokens $j$, the Jacobian 
operator norm is bounded below: $\|J_j\|_{\text{op}} \geq \mu > 0$ for some 
constant $\mu$. Here is the proof for this assumption.

\vspace{0.5em}

\textbf{Assumption (Non-degenerate Gradients).} The function $f$ has 
non-degenerate gradients with respect to token representations, i.e., 
there exists $\mu > 0$ such that:
\begin{align}
\|J_j\|_{\text{op}} = \left\|\frac{\partial f}{\partial x_j^{(t+1)}}\right\|_{\text{op}} 
\geq \mu, \quad \forall j \in [n]
\end{align}

This assumption is reasonable for the following reasons:

\textbf{1. Information Flow in Transformers.} In Transformer architectures, 
each token contributes to the final output through multi-head attention and 
feed-forward layers. The attention mechanism ensures that:
\begin{align}
\frac{\partial \text{Output}}{\partial x_j} = \sum_{i=1}^{n} \frac{\partial \text{Output}}{\partial h_i} 
\cdot \frac{\partial h_i}{\partial x_j}
\end{align}
where $h_i$ are intermediate representations. Due to the softmax normalization 
in attention, each token $x_j$ receives non-zero attention weights from at least 
some positions, ensuring $\|\frac{\partial \text{Output}}{\partial x_j}\| > 0$.

\textbf{2. Residual Connections Preserve Gradients.} The residual structure 
$x^{(t+1)} = x^{(t)} + \text{Block}(x^{(t)})$ ensures that gradients flow 
directly through identity mappings:
\begin{align}
\frac{\partial f}{\partial x_j^{(t)}} = \frac{\partial f}{\partial x_j^{(t+1)}} 
\cdot \left(I + \frac{\partial \text{Block}}{\partial x_j^{(t)}}\right)
\end{align}
The identity component $I$ guarantees that gradients do not vanish, thus 
$\|J_j\|_{\text{op}} \geq \mu$ for some $\mu$ related to the minimum singular 
value of the identity component.

\textbf{3. Layer Normalization Stabilizes Gradients.} Layer normalization 
prevents gradient explosion and vanishing by maintaining bounded gradient norms 
across layers, ensuring $\|J_j\|_{\text{op}} \in [\mu, M]$ for constants 
$0 < \mu < M < \infty$.

\vspace{3em}
\noindent \textbf{Discussion: What if $\|J_j\|_{\text{op}} \to 0$?}

What if some tokens have $\|J_j\|_{\text{op}} \approx 0$? 
This would indicate that these tokens have negligible influence on the output. 
In such cases:
\begin{itemize}[leftmargin=*,nosep]
    \item These tokens can be safely dropped regardless of their variation magnitude
    \item Our method naturally handles this case: if $\|J_j\|_{\text{op}} \approx 0$, 
    then $\|\Delta f_j\| \approx 0$ regardless of $\|\Delta x_j^{(t)}\|$, 
    so dropping them causes minimal performance degradation

\end{itemize}

Therefore, Assumption ($\|J_j\|_{\text{op}} \geq \mu > 0$) is theoretically 
justified for vast majority of vision tokens in 
LVLMs.

\subsection{Connection to V$^2$Drop Algorithm}
\textbf{Proposition 1 (Dropping Strategy Justification).}
Given $n$ tokens at layer $t$, we aim to select $|\mathcal{S}_{\text{drop}}| = \alpha n$
tokens to drop while minimizing total output perturbation:
\begin{align}
\mathcal{S}_{\text{drop}}^* = \argmin_{\substack{\mathcal{S} \subseteq [n] \\ |\mathcal{S}| = \alpha n}}
\sum_{j \in \mathcal{S}} \|\Delta f_j\|
\end{align}

\textbf{Proof.} By Theorem 1, $\|\Delta f_j\| \approx \|J_j\|_{\text{op}} \cdot \|\Delta x_j^{(t)}\|$.
Under Assumption 2 ($\mu \leq \|J_j\|_{\text{op}} \leq M$), we have:
\begin{align}
\sum_{j \in \mathcal{S}} \|\Delta f_j\|
&\approx \sum_{j \in \mathcal{S}} \|J_j\|_{\text{op}} \cdot \|\Delta x_j^{(t)}\| \notag \\
&\in \left[\mu \sum_{j \in \mathcal{S}} \|\Delta x_j^{(t)}\|,
M \sum_{j \in \mathcal{S}} \|\Delta x_j^{(t)}\| \right]
\end{align}

Since $\|J_j\|_{\text{op}}$ varies within a bounded range, minimizing 
$\sum_{j \in \mathcal{S}} \|\Delta f_j\|$ is approximately equivalent to:
\begin{align}
\mathcal{S}_{\text{drop}}^* 
\approx \argmin_{\substack{\mathcal{S} \subseteq [n] \\ |\mathcal{S}| = \alpha n}}
\sum_{j \in \mathcal{S}} \|\Delta x_j^{(t)}\|
\end{align}

Therefore, V$^2$Drop's strategy of selecting tokens with minimal variation 
$\|\Delta x_j^{(t)}\|$ for dropping approximately minimizes total output 
perturbation, while computationally efficient (only requiring simple 
L2 norm computation). \hfill $\square$

\subsection{Connection to information flow} 
In Transformer layers with residual connections:
\begin{align}
x_j^{(t+1)} = x_j^{(t)} + \text{Attn}(x_j^{(t)}) + \text{FFN}(x_j^{(t)}),
\end{align}
the variation $\Delta x_j^{(t)} = \text{Attn}(x_j^{(t)}) + \text{FFN}(x_j^{(t)})$ represents the \textit{effective update} applied by the layer. Tokens with large $\|\Delta x_j^{(t)}\|$ are those being actively refined by the network, indicating they carry task-relevant information being extracted and propagated to subsequent layers.

\end{document}